\documentclass[11pt,twoside]{article} 

\RequirePackage[OT1]{fontenc}
\RequirePackage{amsthm,amsmath,amsfonts,amssymb}
\RequirePackage[colorlinks]{hyperref}
\RequirePackage{hypernat}
\RequirePackage{epsfig, graphicx, color}
\RequirePackage{times}
\RequirePackage{latexsym}
\RequirePackage{psfrag}
\RequirePackage{color}

\usepackage{fullpage}
\usepackage{epsf,epsfig,subfigure}
\usepackage{fancyheadings}
\usepackage{graphics,graphicx}
\usepackage{enumerate}

\theoremstyle{definition}

\theoremstyle{remark}

\newtheorem{theo}{Theorem}[section]

\newtheorem{lem}{Lemma}[section]
\newtheorem{prop}{Proposition}[section]
\newtheorem{cor}{Corollary}[section]

\newtheorem{nota}{Notation}[section]
\newtheorem{de}{Definition}[section]
\newtheorem{exa}{Example}[section]
\newtheorem{as}{Assumption}[section]
\newtheorem{alg}{Algorithm}[section]

\newcommand{\btheo}{\begin{theo}}
\newcommand{\bde}{\begin{de}}
\newcommand{\ble}{\begin{lem}}
\newcommand{\bpr}{\begin{prop}}
\newcommand{\bno}{\begin{nota}}
\newcommand{\bex}{\begin{exa}}
\newcommand{\bcor}{\begin{cor}}
\newcommand{\spro}{\begin{proof}}
\newcommand{\bas}{\begin{as}}
\newcommand{\balg}{\begin{alg}}

\newcommand{\etheo}{\end{theo}}
\newcommand{\ede}{\end{de}}
\newcommand{\ele}{\end{lem}}
\newcommand{\epr}{\end{prop}}
\newcommand{\eno}{\end{nota}}
\newcommand{\eex}{\end{exa}}
\newcommand{\ecor}{\end{cor}}
\newcommand{\fpro}{\end{proof}}
\newcommand{\eas}{\end{as}}
\newcommand{\ealg}{\end{alg}}

\newtheorem{theos}{Theorem}
\newtheorem{props}{Proposition}
\newtheorem{lems}{Lemma}
\newtheorem{cors}{Corollary}

\newtheorem{exas}{Example}
\newtheorem{algs}{Algorithm}
\newtheorem{asss}{Asumption}
\newtheorem{defns}{Definition}

\newcommand{\btheos}{\begin{theos}}
\newcommand{\etheos}{\end{theos}}
\newcommand{\bprops}{\begin{props}}
\newcommand{\eprops}{\end{props}}
\newcommand{\bdes}{\begin{defns}}
\newcommand{\edes}{\end{defns}}
\newcommand{\blems}{\begin{lems}}
\newcommand{\elems}{\end{lems}}
\newcommand{\bcors}{\begin{cors}}
\newcommand{\ecors}{\end{cors}}
\newcommand{\bexs}{\begin{exas}}
\newcommand{\eexs}{\end{exas}}
\newcommand{\balgs}{\begin{algs}}
\newcommand{\ealgs}{\end{algs}}
\newcommand{\bass}{\begin{asss}}
\newcommand{\eass}{\end{asss}}

\begin{document}
	\begin{center}
	{\bf{\LARGE{A statistical perspective on randomized sketching \\ for ordinary least-squares}}}

	\vspace*{.1in}
	\begin{tabular}{cc}
	Garvesh Raskutti$^1$ & Michael Mahoney $^{2,3}$ \\
	\end{tabular}

	\vspace*{.1in}

	\begin{tabular}{c}
	  $^1$ Department of Statistics \& Department of Computer Science, University of Wisconsin Madison \\
	$^2$ International Computer Science Institute\\
	$^3$ Department of Statistics, University of California Berkeley
	\end{tabular}

	\vspace*{.1in}


	\end{center}

\begin{abstract}
\noindent
We consider statistical as well as algorithmic aspects of solving large-scale least-squares (LS) problems using randomized sketching algorithms.
For a LS problem with input data $(X, Y) \in \mathbb{R}^{n \times p} \times \mathbb{R}^n$, sketching algorithms use a ``sketching matrix,'' $S\in\mathbb{R}^{r \times n}$, where $r \ll n$. Then, rather than solving the LS problem using the full data $(X,Y)$, sketching algorithms solve the LS problem using only the ``sketched data'' $(SX, SY)$. Prior work has typically adopted an \emph{algorithmic perspective}, in that it has made no statistical assumptions on the input $X$ and $Y$, and instead it has been assumed that the data $(X,Y)$ are fixed and worst-case (WC). 
Prior results show that, when using sketching matrices such as random projections and leverage-score sampling algorithms, with $p \lesssim r \ll n$, the WC error is the same as solving the original problem, up to a small constant. 
From a \emph{statistical perspective}, we typically consider the mean-squared error performance of randomized sketching algorithms, when data $(X, Y)$ are generated according to a statistical linear model $Y = X \beta + \epsilon$, where $\epsilon$ is a noise process. In this paper, we provide a rigorous comparison of both perspectives leading to insights on how they differ. To do this, we first develop a framework for assessing, in a unified manner, algorithmic and statistical aspects of randomized sketching 
methods. We then consider the statistical prediction efficiency (PE) and the statistical residual efficiency (RE) of the sketched LS estimator; and 
we use our framework to provide upper bounds for several types of random projection and random sampling sketching algorithms. Among other results, we show that the RE can be upper bounded when $p \lesssim r \ll n$ while the PE typically requires the sample size $r$ to be substantially larger. Lower bounds developed in subsequent results show that our upper bounds on PE can not be improved.%
\end{abstract}


\section{Introduction}

Recent work in large-scale data analysis has focused on developing so-called 
sketching algorithms: given a data set and an objective function of 
interest, construct a small ``sketch'' of the full data set, e.g., by using 
random sampling or random projection methods, and use that sketch as a 
surrogate to perform computations of interest for the full data 
set (see~\cite{Mah-mat-rev_BOOK} for a review).
Most effort in this area has adopted an \emph{algorithmic perspective}, 
whereby one shows that, when the sketches are constructed appropriately, 
one can obtain answers that are approximately as good as the exact answer
for the input data at hand, in less time than would be required to compute 
an exact answer for the data at hand.
In statistics, however, one is often more interested in how well a procedure
performs relative to an hypothesized model than how well it performs on the 
particular data set at hand. Thus an important to question to consider is whether
the insights from the algorithmic perspective of sketching carry over to the statistical 
setting. 

Thus, in this paper, we develop a unified approach that considers both the 
 \emph{statistical perspective} as well as \emph{algorithmic perspective} on 
recently-developed randomized sketching algorithms, and we provide bounds on 
two statistical objectives for several types of random projection 
and random sampling sketching algorithms.

\subsection{Overview of the problem}

The problem we consider in this paper is the ordinary least-squares (LS or 
OLS) problem:
given as input a matrix $X \in \mathbb{R}^{n \times p}$ of observed features 
or covariates and a vector $Y \in \mathbb{R}^n$ of observed responses,  
return as output a vector $\beta_{OLS}$ that solves the following 
optimization problem:

\begin{equation}
\label{NoiseLinMod}
\beta_{OLS} = \arg \min_{\beta\in\mathbb{R}^p}\|Y - X \beta\|_2^2  .
\end{equation}

\noindent
We will assume that $n$ and $p$ are both very large, with $n \gg p$, and 
for simplicity we will assume $\mbox{rank}(X) = p$, e.g., to ensure a unique 
full-dimensional solution. 
The OLS solution, $\beta_{OLS} = (X^T X)^{-1} X^T Y$, has a number of 
well-known desirable statistical properties~\cite{ChatterjeeHadi88}; and 
it is also well-known that the running time or computational complexity for 
this problem is $O(n p^2)$~\cite{GVL96}.%
\footnote{That is, $O(n p^2)$ time suffices to compute the LS solution 
from Problem~(\ref{NoiseLinMod}) for arbitrary or worst-case input, with, 
e.g., the Cholesky Decomposition on the normal equations, with a QR 
decomposition, or with the Singular Value Decomposition~\cite{GVL96}.}
For many modern applications, however, $n$ may be on the order of 
$10^6-10^9$ and $p$ may be on the order of $10^3-10^4$, and thus computing 
the exact LS solution with traditional $O(n p^2)$ methods can be 
computationally challenging. 
This, coupled with the observation that approximate answers often suffice 
for downstream applications, has led to a large body of work on developing 
fast approximation algorithms to the LS problem~\cite{Mah-mat-rev_BOOK}.

One very popular approach to reducing computation is to perform LS on a 
carefully-constructed ``sketch'' of the full data set. 
That is, rather than computing a LS estimator from 
Problem~(\ref{NoiseLinMod}) from the full data $(X,Y)$, generate ``sketched 
data'' $(SX, SY)$ where $S \in \mathbb{R}^{r \times n}$, with $r \ll n$, is a 
``sketching matrix,'' and then compute a LS estimator from the following 
sketched problem: 

\begin{equation}
\label{NoiseLinModSketched}
\beta_S \in \arg \min_{\beta \in \mathbb{R}^p}\|SY - S X \beta\|_2^2. 
\end{equation}

\noindent
Once the sketching operation has been performed, the additional computational 
complexity of $\beta_S$ is $O(r p^2)$, i.e., simply call a 
traditional LS solver on the sketched problem. 
Thus, when using a sketching algorithm, two criteria are important:
first, ensure the accuracy of the sketched LS estimator is comparable to,
e.g., not much worse, than the performance of the original LS estimator; and 
second, ensure that computing and applying the sketching matrix $S$ is not 
too computationally intensive, e.g., that is faster than solving the 
original problem exactly.

\subsection{Prior results}

Random sampling and random projections provide two approaches to construct 
sketching matrices $S$ that satisfy both of these criteria and that have 
received attention recently in the computer science community. 
Very loosely speaking, a random projection matrix $S$ is a dense matrix,%
\footnote{The reader should, however, be aware of recently-developed 
input-sparsity time random projection methods~\cite{CW13sparse_STOC,MM13_STOC,NH13}.}   
where each entry is a mean-zero bounded-variance Gaussian or 
Rademacher random variable, although other constructions based on 
randomized Hadamard transformations are also of interest; and 
a random sampling matrix $S$ is a very sparse matrix that has exactly $1$ 
non-zero entry (which typically equals one multiplied by a rescaling factor) 
in each row, where that one non-zero can be chosen uniformly, non-uniformly 
based on hypotheses about the data, or non-uniformly based on empirical 
statistics of the data such as the leverage scores of the matrix $X$.
In particular, note that a sketch constructed from an $r \times n$ random 
projection matrix $S$ consists of $r$ linear combinations of most or all of 
the rows of $(X,Y)$, and a sketch constructed from a random sampling matrix 
$S$ consists of $r$ typically-rescaled rows of $(X,Y)$.
Random projection algorithms have received a great deal of attention more 
generally, largely due to their connections with the Johnson-Lindenstrauss 
lemma~\cite{JohnsonLindenstrauss} and its extensions; and random sampling 
algorithms have received a great deal of attention, largely due to their
applications in large-scale data analysis applications~\cite{DrinMah09}.
A detailed overview of random projection and random sampling algorithms for 
matrix problems may be found in the recent monograph of~\cite{Mah-mat-rev_BOOK}.
Here, we briefly summarize the most relevant aspects of the~theory.

In terms of running time guarantees, the running time bottleneck for random 
projection algorithms for the LS problem is the application of the 
projection to the input data, i.e., actually performing the matrix-matrix
multiplication to implement the projection and compute the sketch. 
By using fast Hadamard-based random projections, however,~\cite{DrinMuthuMahSarlos11} developed a random 
projection algorithm that runs on arbitrary or worst-case input 
in $o(np^2)$ time.
(See~\cite{DrinMuthuMahSarlos11} for a precise statement of the running 
time.)
As for random sampling, it is trivial to implement uniform random sampling, 
but it is very easy to show examples of input data on which uniform sampling 
performs very poorly.
On the other hand, ~\cite{DMM06,DMMW12_JMLR} have shown that 
if the random sampling is performed with respect to nonuniform importance
sampling probabilities that depend on the \emph{empirical statistical 
leverage scores} of the input matrix $X$, i.e., the diagonal entries of the 
\emph{hat matrix} $H = X(X^T X)^{-1} X^T$, then one obtains a random 
sampling algorithm that achieves much better results for arbitrary or 
worst-case input.%

Leverage scores have a long history in robust statistics and experimental design. In the robust statistics community, samples with high leverage scores are typically flagged as potential 
outliers (see, e.g.,~\cite{ChatterjeeHadi86,ChatterjeeHadi88, Hampel86, HW78, Huber81}). In the experimental design community, samples with high leverage have been shown to improve overall efficiency, provided that the underlying statistical model is accurate (see, e.g.,~\cite{Royall70, Zaslavsky08}). This should be contrasted with their use in theoretical computer science. From the algorithmic perspective of worst-case analysis, that was adopted by ~\cite{DrinMuthuMahSarlos11} and~\cite{DMMW12_JMLR}, samples with high leverage tend to contain the most important information for subsampling/sketching, and 
thus it is beneficial for worst-case analysis to bias the random sample to 
include samples with large statistical leverage scores or to rotate to a random basis where the leverage scores are approximately uniformized. 

The running-time bottleneck for this leverage-based random sampling 
algorithm is the computation of the leverage scores of the input data; and
the obvious well-known algorithm for this involves $O(np^2)$ time to perform a QR 
decomposition to compute an orthogonal basis for $X$~\cite{GVL96}. 
By using fast Hadamard-based random projections, however, ~\cite{DMMW12_JMLR} showed that one can compute approximate QR 
decompositions and thus approximate leverage scores in $o(np^2)$ time, and
(based on previous work~\cite{DMM06}) this immediately implies a 
leverage-based random sampling algorithm that runs on arbitrary or 
worst-case input in $o(np^2)$ time~\cite{DMMW12_JMLR}. 
Readers interested in the practical performance of these randomized 
algorithms should consult \textsc{Bendenpik}~\cite{AMT10} or 
\textsc{LSRN}~\cite{MSM14_SISC}.

In terms of accuracy guarantees, both \cite{DrinMuthuMahSarlos11} and
\cite{DMMW12_JMLR} prove that their respective random 
projection and leverage-based random sampling LS sketching algorithms each 
achieve the following worst-case (WC) error guarantee:
for any arbitrary $(X, Y)$,
\begin{equation}
\label{eqn:ErrorWCE}
\|Y - X \beta_S \|_2^2 \leq (1+ \kappa)\|Y - X \beta_{OLS} \|_2^2,
\end{equation} 
with high probability for some pre-specified error parameter $\kappa \in (0,1)$.%
\footnote{The quantity $\|\beta_S-\beta_{OLS}\|_2^2$ is also bounded by \cite{DrinMuthuMahSarlos11} 
and \cite{DMMW12_JMLR}.}
This $1+ \kappa$ relative-error guarantee%
\footnote{The nonstandard parameter $\kappa$ is used here for the error 
parameter since $\epsilon$ is used below to refer to the noise or error 
process.}
is extremely strong, and it is applicable to arbitrary or worst-case input.
That is, whereas in statistics one typically assumes a model, e.g., a 
standard linear model on $Y$,
\begin{equation}
\label{EqnLinModel}
Y = X \beta + \epsilon,
\end{equation}
where $\beta \in \mathbb{R}^p$ is the true parameter and 
$\epsilon \in \mathbb{R}^n$ is a standardized noise vector, with 
$\mathbb{E}[\epsilon]=0$ and $\mathbb{E}[\epsilon\epsilon^T]=I_{n \times n}$, 
in \cite{DrinMuthuMahSarlos11} 
and \cite{DMMW12_JMLR}
no statistical model is assumed on $X$ and $Y$, and 
thus the running time and quality-of-approximation 
bounds apply to any arbitrary $(X,Y)$ input data. 

\subsection{Our approach and main results}

In this paper, we adopt a statistical perspective on these
randomized sketching algorithms, and we address the following fundamental
questions.
First, under a standard linear model, e.g., as given in 
Eqn.~(\ref{EqnLinModel}), what properties of a sketching matrix $S$ are 
sufficient to ensure low statistical error, e.g., mean-squared, error?
Second, how do existing random projection algorithms and leverage-based 
random sampling algorithms perform by this statistical measure? 
Third, how does this relate to the properties of a sketching matrix $S$ that
are sufficient to ensure low worst-case error, e.g., of the form of
Eqn.~(\ref{eqn:ErrorWCE}), as has been established 
previously in \cite{DrinMuthuMahSarlos11,DMMW12_JMLR,Mah-mat-rev_BOOK}?
We address these related questions in a number of steps. 

In Section~\ref{SecFramework}, we will present a framework for evaluating
the algorithmic and statistical properties of randomized 
sketching methods in a unified manner; and we will show that providing 
worst-case error bounds of the form of Eqn.~(\ref{eqn:ErrorWCE}) and 
providing bounds on two related statistical objectives boil down to 
controlling different structural properties of how the sketching matrix $S$ 
interacts with the left singular subspace of the design matrix.
In particular, we will consider the oblique projection matrix,
$\Pi_S^U = U (SU)^{\dagger} S$, where $(\cdot)^\dagger$ denotes the 
Moore-Penrose pseudo-inverse of a matrix and $U$ is the left singular matrix 
of $X$. 
This framework will allow us to draw a comparison between the worst-case error
and two related statistical efficiency criteria, the statistical 
prediction efficiency (PE) (which is based on the prediction error 
$\mathbb{E}[\|X(\widehat{\beta} - \beta)\|_2^2]$ and which is given in 
Eqn.~(\ref{DefnSPE}) below) and the statistical residual efficiency (RE)
(which is based on residual error $\mathbb{E}[\|Y - X \widehat{\beta}\|_2^2]$ and 
which is given in Eqn.~(\ref{DefnSRE}) below); and 
it will allow us to provide sufficient conditions that any sketching matrix 
$S$ must satisfy in order to achieve performance guarantees for these two 
statistical objectives. 

In Section~\ref{SecMainResults}, we will present our main theoretical 
results, which consist of bounds for these two statistical quantities for 
variants of random sampling and random projection sketching algorithms.
In particular, we provide upper bounds on the PE and RE (as well as the 
worst-case WC) for four sketching schemes: 
(1) an approximate leverage-based random sampling algorithm, as is 
analyzed by \cite{DMMW12_JMLR}; 
(2) a variant of leverage-based random sampling, where the random samples
are \emph{not} re-scaled prior to their inclusion in the sketch, as is 
considered by \cite{MMY14_ICML,MMY15}; 
(3) a vanilla random projection algorithm, where $S$ is a random matrix 
containing i.i.d. Gaussian or Rademacher random variables, as is popular in 
statistics and scientific computing; and 
(4) a random projection algorithm, where $S$ is a random 
Hadamard-based random projection, as analyzed in \cite{BoutsGitt12}. 
For sketching schemes (1), (3), and (4), our upper bounds for each of the two 
measures of statistical efficiency are identical up to constants; and
they show that the RE scales as $1+ \frac{p}{r}$, while the PE scales as 
$\frac{n}{r}$. 
In particular, this means that it is possible to obtain good bounds for the 
RE when $p \lesssim r \ll n$ (in a manner similar to the sampling 
complexity of the WC bounds); but in order to obtain even near-constant 
bounds for PE, $r$ must be at least of constant order compared to $n$.
We then present a lower bound developed in subsequent work by \cite{PilanciWainwright} which shows that under general conditions
on $S$, our upper bound of $\frac{n}{r}$ for PE can not be improved. 
For the sketching scheme (2), we show, on the other hand, that under the 
strong assumption that there are $k$ ``large'' leverage scores and the 
remaining $n-k$ are ``small,'' then the WC scales as $1+ \frac{p}{r}$, the 
RE scales as $1+ \frac{pk}{rn}$, and the PE scales as $\frac{k}{r}$.
That is, sharper bounds are possible for leverage-score sampling without 
re-scaling in the statistical setting, but much stronger assumptions are 
needed on the input~data. 

In Section~\ref{SecSimulations}, we will supplement our theoretical results 
by presenting our main empirical results, which consist of an evaluation of 
the complementary properties of random sampling versus random projection 
methods. Our empirical results support our theoretical results, and they also show that 
for $r$ larger than $p$ but much closer to $p$ than $n$, projection-based methods tend to 
out-perform sampling-based methods, while for $r$ significantly larger than 
$p$, our leverage-based sampling methods perform slightly better. 
In Section~\ref{SecDiscussion}, we will provide a brief discussion and 
conclusion and we provide proofs of our main results in the Appendix.

\subsection{Additional related work}

Very recently \cite{MMY14_ICML} considered 
statistical aspects of leverage-based sampling algorithms (called 
\emph{algorithmic leveraging} in \cite{MMY14_ICML}).
Assuming a standard linear model on $Y$ of the form of 
Eqn.~\eqref{EqnLinModel}, the authors developed first-order Taylor 
approximations to the statistical relative efficiency of different 
estimators computed with leverage-based sampling algorithms, and they 
verified the quality of those approximations with computations on real and 
synthetic data. 
Taken as a whole, their results suggest that, if one is interested in the
statistical performance of these randomized sketching algorithms, then there 
are nontrivial trade-offs that are not taken into account by standard
worst-case analysis.
Their approach, however, does not immediately apply to random projections or other more 
general sketching matrices. Further, the realm of applicability of the first-order Taylor approximation was not precisely quantified, and they left open the question of structural characterizations of random sketching matrices that were sufficient to ensure good statistical properties on the sketched data. We address these issues in this paper.

After the appearance of the original technical report version of this paper \cite{RaskuttiMahoney}, we were made aware of subsequent work by \cite{PilanciWainwright}, who also consider a statistical perspective on sketching. Amongst other results, they develop a lower bound which confirms that using a single randomized sketching matrix $S$ can not achieve a PE better than $\frac{n}{r}$. This lower bound complements our upper bounds developed in this paper. Their main focus is to use this insight to develop an iterative sketching scheme which yields bounds on the PE when an $ r\times n$ sketch is applied repeatedly.

\section{General framework and structural results}
\label{SecFramework}

In this section, we develop a framework that allows us to view the
algorithmic and statistical perspectives on LS problems from a 
common perspective. We then use this framework to show that existing worst-case bounds as well as our novel statistical bounds for the mean-squared errors can be expressed in terms of different structural conditions on how the sketching matrix $S$ interacts with the data $(X,Y)$. 

\subsection{A statistical-algorithmic framework}

Recall that we are given as input a data set, 
$(X, Y) \in \mathbb{R}^{n\times p} \times \mathbb{R}^n$, and the objective 
function of interest is the standard LS objective,
as given in Eqn.~(\ref{NoiseLinMod}).
Since we are assuming, without loss of generality, that 
$\mbox{rank}(X)=p$, we have that 
\begin{equation}
\label{eqn:beta_opt_full}
\beta_{OLS} = X^{\dagger}Y = (X^T X)^{-1}X^T Y, 
\end{equation}
where $(\cdot)^{\dagger}$ denotes the Moore-Penrose pseudo-inverse of a matrix, and where the second equality follows since $\mbox{rank}(X)=p$

To present our framework and objectives, let $S \in \mathbb{R}^{r \times n}$ 
denote an \emph{arbitrary} sketching matrix. 
That is, although we will be most interested in sketches constructed from 
random sampling or random projection operations, for now we let $S$ be 
\emph{any} $r \times n$ matrix.
Then, we are interested in analyzing the performance of objectives 
characterizing the quality of a ``sketched'' LS objective, as given in 
Eqn~(\ref{NoiseLinModSketched}), where again we are interested in solutions of the form 
\begin{equation}
\label{eqn:beta_opt_sketched}
\beta_S=(SX)^{\dagger}SY . 
\end{equation}
(We emphasize that this does \emph{not} in general equal 
$((SX)^T SX)^{-1}(SX)^T SY$, since the inverse will \emph{not} exist if the 
sketching process does not preserve rank.)
Our goal here is to compare the performance of $\beta_S$ to
$\beta_{OLS}$.
We will do so by considering three related performance criteria, two of a statistical flavor, and one of a more algorithmic or worst-case flavor.

From a statistical perspective, it is common to assume a standard linear 
model on $Y$, 
\begin{equation*}
Y = X \beta + \epsilon,
\end{equation*}
where we remind the reader that $\beta \in \mathbb{R}^p$ is the true 
parameter and $\epsilon \in \mathbb{R}^n$ is a standardized noise 
vector, with $\mathbb{E}[\epsilon]=0$ and 
$\mathbb{E}[\epsilon\epsilon^T]=I_{n \times n}$. 
From this statistical perspective, we will consider the following two~criteria.
\begin{itemize}
\item
The first statistical criterion we consider is the
\emph{prediction efficiency} (PE), defined as follows:
\begin{equation}
\label{DefnSPE}
C_{PE}(S) = \frac{\mathbb{E}[\|X (\beta - \beta_S)\|_2^2]}{\mathbb{E}[\|X (\beta - \beta_{OLS})\|_2^2]}  , 
\end{equation}
where the expectation $\mathbb{E}[\cdot]$ is taken over the random noise 
$\epsilon$. 
\item
The second statistical criterion we consider is the 
\emph{residual efficiency} (RE), defined as follows:
\begin{equation}
\label{DefnSRE}
C_{RE}(S) = \frac{\mathbb{E}[\|Y - X \beta_S\|_2^2]}{\mathbb{E}[\|Y - X \beta_{OLS}\|_2^2]}  , 
\end{equation}
where, again, the expectation $\mathbb{E}[\cdot]$ is taken over the random 
noise $\epsilon$.
\end{itemize}
Recall that the standard relative statistical efficiency for two estimators 
$\beta_1$ and $\beta_2$ is defined as 
$\mbox{eff}(\beta_1,\beta_2)=\frac{\mbox{Var}(\beta_1)}{\mbox{Var}(\beta_2)}$, 
where $\mbox{Var}(\cdot)$ denotes the variance of the 
estimator (see e.g., \cite{Lehmann98}). 
For the PE, we have replaced the variance of each estimator by the 
mean-squared prediction error. 
For the RE, we use the term residual since for any estimator 
$\widehat{\beta}$, $Y - X \widehat{\beta}$ are the residuals for estimating 
$Y$. 

From an algorithmic perspective, there is no noise process $\epsilon$.
Instead, $X$ and $Y$ are arbitrary, and $\beta$ is simply computed 
from Eqn~(\ref{eqn:beta_opt_full}).
To draw a parallel with the usual statistical generative process, however, 
and to understand better the relationship between various objectives, 
consider ``defining'' $Y$ in terms of $X$ by the following ``linear model'':
\begin{equation*}
Y = X \beta + \epsilon,
\end{equation*}
where $\beta \in \mathbb{R}^p$ and $\epsilon \in \mathbb{R}^n$.  
Importantly, $\beta$ and $\epsilon$ here represent different quantities than 
in the usual statistical setting. 
Rather than $\epsilon$ representing a noise process and $\beta$ 
representing a ``true parameter'' that is observed through a noisy $Y$, 
here in the algorithmic setting, we will take advantage of the 
rank-nullity theorem in linear algebra to relate $X$ and $Y$.%
\footnote{The rank-nullity theorem asserts that given any 
matrix $X \in \mathbb{R}^{n \times p}$ and vector $Y \in \mathbb{R}^n$, 
there exists a unique decomposition $Y = X \beta + \epsilon$, where $\beta$ 
is the projection of $Y$ on to the range space of $X^T$ and 
$\epsilon = Y-X\beta$ lies in the null-space of $X^T$~\cite{Meyer00}.}   
To define a ``worst case model'' $ Y = X \beta + \epsilon$ for the 
algorithmic setting, one can view the ``noise'' process $\epsilon$ to 
consist of any vector that lies in the null-space of $X^T$.
Then, since the choice of $\beta \in \mathbb{R}^p$ is arbitrary, one can 
construct any arbitrary or worst-case input data $Y$.
From this algorithmic case, we will consider the following~criterion.
\begin{itemize}
\item
The algorithmic criterion we consider is the \emph{worst-case} (WC) error, 
defined as follows:
\begin{equation}
\label{DefnWCE}
C_{WC}(S) = \sup_{Y} \frac{\|Y - X \beta_S\|_2^2}{\|Y - X \beta_{OLS}\|_2^2}.
\end{equation}
\end{itemize}
This criterion is worst-case since we take a supremum $Y$, 
and it is the performance criterion that is analyzed in
\cite{DrinMuthuMahSarlos11} and
\cite{DMMW12_JMLR},
as bounded in Eqn.~(\ref{eqn:ErrorWCE}).

Writing $Y$ as $X \beta + \epsilon$, where $X^T \epsilon = 0$, the WC error can be re-expressed as:
\begin{equation*}
C_{WC}(S) = \sup_{Y= X \beta + \epsilon,\; X^T \epsilon = 0} \frac{\|Y - X \beta_S\|_2^2}{\|Y - X \beta_{OLS}\|_2^2}.
\end{equation*}
Hence, in the worst-case algorithmic setup, we take a supremum over $\epsilon$, where $X^T \epsilon = 0$, whereas in the statistical setup, we take an expectation over $\epsilon$ where $\mathbb{E}[\epsilon] = 0$.

Before proceeding, several other comments about this algorithmic-statistical 
framework and our objectives are worth mentioning.
\begin{itemize}
\item
From the perspective of our two linear models, we have that 
$\beta_{OLS} = \beta + (X^T X)^{-1} X^T \epsilon$. 
In the statistical setting, since 
$\mathbb{E}[\epsilon \epsilon^T] = I_{n \times n}$, it follows that 
$\beta_{OLS}$ is a random variable with $\mathbb{E}[\beta_{OLS}] = \beta$ 
and $\mathbb{E}[(\beta - \beta_{OLS})(\beta - \beta_{OLS})^T] = (X^T X)^{-1}$.
In the algorithmic setting, on the other hand, since $X^T \epsilon = 0$, it follows that $\beta_{OLS} = \beta$. 
\item
$C_{RE}(S)$ is a statistical analogue of the worst-case algorithmic 
objective $C_{WC}(S)$, since both consider the ratio of the metrics 
$\frac{\|Y - X \beta_S\|_2^2}{\|Y - X \beta_{OLS}\|_2^2}$.
The difference is that a $\sup$ over $Y$ in the algorithmic setting is 
replaced by an expectation over noise $\epsilon$ in the statistical setting. 
A natural question is whether there is an algorithmic analogue of $C_{PE}(S)$. 
Such a performance metric would be:
\begin{equation}
\label{eqn:nonexistent_obj}
\sup_{Y} \frac{\|X (\beta - \beta_S)\|_2^2}{\|X (\beta-\beta_{OLS})\|_2^2},
\end{equation}
where $\beta$ is the projection of $Y$ on to the range space of $X^T$.  
However, since $\beta_{OLS} = \beta + (X^T X)^{-1} X^T \epsilon$ and since $X^T \epsilon = 0$, $\beta_{OLS} = \beta $ in the 
algorithmic setting, the denominator 
of Eqn.~(\ref{eqn:nonexistent_obj}) equals zero, and thus the objective in 
Eqn.~(\ref{eqn:nonexistent_obj}) is not well-defined.
The ``difficulty'' of computing or approximating this objective parallels 
our results below that show that approximating $C_{PE}(S)$ is much more 
challenging (in terms of the number of samples needed) than approximating 
$C_{RE}(S)$. 
\item
In the algorithmic setting, the sketching matrix $S$ and the objective
$C_{WC}(S)$ can depend on $X$ and $Y$ in any arbitrary way, but in the 
following we consider only sketching matrices that are either independent 
of both $X$ and $Y$ or depend only on $X$ (e.g., via the statistical 
leverage scores of $X$). 
In the statistical setting, $S$ is allowed to depend on $X$, but not on 
$Y$, as any dependence of $S$ on $Y$ might introduce correlation between 
the sketching matrix and the noise variable $\epsilon$.
Removing this restriction is of interest, especially since one can obtain WC 
bounds of the form Eqn.~(\ref{eqn:ErrorWCE}) by constructing $S$ by randomly 
sampling according to an importance sampling distribution that depends on 
the \emph{influence scores}---essentially the leverage scores of the matrix $X$ 
augmented with $-Y$ as an additional column---of the $(X, Y)$ pair.
\item
Both $C_{PE}(S)$ and $C_{RE}(S)$ are qualitatively related to quantities 
analyzed by \cite{MMY14_ICML,MMY15}. 
In addition, $C_{WC}(S)$ is qualitatively similar to 
$\mbox{Cov}(\widehat{\beta} | Y)$ in \cite{MMY14_ICML,MMY15}, since in the algorithmic setting $Y$ is 
treated as fixed; and $C_{RE}(S)$ is qualitatively similar to 
$\mbox{Cov}(\widehat{\beta})$ in \cite{MMY14_ICML,MMY15}, since in the statistical setting $Y$ is 
treated as random and coming from a linear model. 
That being said, the metrics and results we present in this paper are not 
directly comparable to those of \cite{MMY14_ICML,MMY15} since, e.g., they had a slightly different setup 
than we have here, and since they used a first-order Taylor approximation while we 
do not.
\end{itemize}

\subsection{Structural results on sketching matrices}

We are now ready to develop structural conditions characterizing how the 
sketching matrix $S$ interacts with the data matrix $X$ that will allow us 
to provide upper bounds for the quantities $C_{WC}(S), C_{PE}(S)$, and 
$C_{RE}(S)$. 
To do so, recall that given the data matrix $X$, we can express the 
singular value decomposition of $X$ as $X = U \Sigma V^T$, where 
$U \in \mathbb{R}^{n \times p}$ is an orthogonal matrix, i.e., 
$U^T U  = I_{p \times p}$. 
In addition, we can define the \emph{oblique projection} matrix
\begin{equation}
\Pi_S^U := U (SU)^\dagger S  .
\end{equation} 
Note that if $\mbox{rank}(SX) = p$, then $\Pi_S^U$ can be expressed as
$\Pi_S^U = U (U^T S^T S U)^{-1} U^T S^T S$, since $U^T S^T S U$ is 
invertible.
Importantly however, depending on the properties of $X$ and how $S$ is 
constructed, it can easily happen that $\mbox{rank}(SX) < p$, even if 
$\mbox{rank}(X) = p$. 

Given this setup, we can now state the following lemma, the proof of which 
may be found in Section~\ref{SecProofs-first-lem}.
This lemma characterizes how $C_{WC}(S)$, $C_{PE}(S)$, and $C_{RE}(S)$ 
depend on different structural properties of $\Pi_S^U$ and $SU$. 

\blems
\label{LemProj}
For the algorithmic setting,
\begin{equation*}
C_{WC}(S) = 1+ \sup_{\delta \in \mathbb{R}^p, U^T \epsilon = 0 } \biggr[ \frac{\| (I_{p \times p} - (SU)^{\dagger}(SU) )\delta\|_2^2}{\|\epsilon\|_2^2} + \frac{\|\Pi_S^U \epsilon \|_2^2}{\|\epsilon\|_2^2}\biggr].
\end{equation*}
For the statistical setting,
\begin{equation*}
C_{PE}(S) = \frac{\| (I
_{p \times p}- (SU)^{\dagger} SU) \Sigma V^T \beta\|_2^2}{p} + \frac{\|\Pi_S^U\|_F^2}{p},
\end{equation*}
and
\begin{equation*}
C_{RE}(S) = 1 + \frac{\| (I
_{p \times p}- (SU)^{\dagger} SU) \Sigma V^T \beta\|_2^2}{n-p} + \frac{\|\Pi_S^U\|_F^2 - p}{n-p} = 1+ \frac{C_{SPE}(S) - 1}{n/p - 1 } .
\end{equation*}
\elems

\noindent
Several points are worth making about Lemma~\ref{LemProj}.
\begin{itemize}
\item
For all $3$ criteria, the term which involves $(SU)^{\dagger} SU$ is a 
``bias'' term that is non-zero in the case that $\mbox{rank}(SU) < p$. 
For $C_{PE}(S)$ and $C_{RE}(S)$, the term corresponds exactly to the 
statistical bias; and if $\mbox{rank}(SU) = p$, meaning that $S$ is a 
\emph{rank-preserving} sketching matrix, then the bias term equals $0$, 
since $(SU)^{\dagger} SU = I_{p \times p}$. 
In practice, if $r$ is chosen smaller than $p$ or larger than but very close 
to $p$, it may happen that $\mbox{rank}(SU) < p$, in which case this bias is 
incurred.
\item
The final equality $C_{RE}(S) = 1+ \frac{C_{PE}(S) - 1}{n/p - 1 }$ shows 
that in general it is much more difficult (in terms of the number of 
samples needed) to obtain bounds on $C_{PE}(S)$ than $C_{RE}(S)$---since 
$C_{RE}(S)$ re-scales $C_{PE}(S)$ by $p/n$, which is much less than $1$. 
This will be reflected in the main results below, where the scaling of 
$C_{RE}(S)$ will be a factor of $p/n$ smaller than $C_{PE}(S)$.  
In general, it is significantly more difficult to bound $C_{PE}(S)$, since 
$\|X(\beta - \beta_{OLS})\|_2^2$ is $p$, whereas $\|Y - X \beta_{OLS}\|_2^2$ 
is $n-p$, and so there is much less margin for error in approximating 
$C_{PE}(S)$. 
\item
In the algorithmic or worst-case setting, 
$\sup_{\epsilon \in \mathbb{R}^n/\{ 0\}, \Pi^U \epsilon = 0 } \frac{\|\Pi_S^U \epsilon \|_2^2}{\|\epsilon\|_2^2}$ 
is the relevant quantity, whereas in the statistical setting 
$\|\Pi_S^U\|_F^2$ is the relevant quantity. 
The Frobenius norm enters in the statistical setting because we are taking 
an average over homoscedastic noise, and so the $\ell_2$ norm of the 
eigenvalues of $\Pi_S^U$ need to be controlled. 
On the other hand, in the algorithmic or worst-case setting, the worst 
direction in the null-space of $U^T$ needs to be controlled, and thus the 
spectral norm enters.
\end{itemize}

\section{Main theoretical results}
\label{SecMainResults}

In this section, we provide upper bounds for $C_{WC}(S)$, $C_{PE}(S)$, 
and $C_{RE}(S)$, where $S$ correspond to random sampling and random 
projection matrices. 
In particular, we provide upper bounds for $4$ sketching matrices:
(1) a vanilla leverage-based random sampling algorithm from \cite{DMMW12_JMLR}; 
(2) a variant of leverage-based random sampling, where the random samples
are \emph{not} re-scaled prior to their inclusion in the sketch;
(3) a vanilla random projection algorithm, where $S$ is a random matrix 
containing i.i.d. sub-Gaussian random variables; and 
(4) a random projection algorithm, where $S$ is a random 
Hadamard-based random projection, as analyzed in \cite{BoutsGitt12}.

\subsection{Random sampling methods}
\label{SecSampling}

Here, we consider random sampling algorithms.
To do so, first define a random sampling matrix $\tilde{S} \in \mathbb{R}^n$ 
as follows: $\tilde{S}_{ij} \in \{0, 1\}$ for all $(i,j)$ and 
$\sum_{j=1}^n \tilde{S}_{ij} = 1$, where each row has an independent 
multinomial distribution with probabilities $(p_i)_{i=1}^n$. 
The matrix of cross-leverage scores is defined as 
$L = U U^T \in \mathbb{R}^{n \times n}$, and $\ell_i = L_{ii}$ denotes the 
leverage score corresponding to the $i^{th}$ sample. 
Note that the leverage scores satisfy 
$\sum_{i=1}^n{\ell_i} = \mbox{trace}(L) = p$ and $0 \leq \ell_i \leq 1$.
 
The sampling probability distribution we consider $(p_i)_{i=1}^n$ is of the 
form $p_i = (1 - \theta) \frac{\ell_i}{p} + \theta q_i$, where 
$\{q_i\}_{i=1}^n$ satisfies $0 \leq q_i \leq 1$ and $\sum_{i=1}^n {q_i} = 1$ is 
an arbitrary probability distribution, and $0 \leq \theta < 1$. In other words, it is a convex combination of a leverage-based distribution and another arbitrary distribution. Note that for $\theta = 0$, the probabilities are proportional to the leverage scores, whereas for $\theta = 1$, the probabilities follow $\{q_i\}_{i=1}^{n}$.

We consider two sampling matrices, one where the random sampling matrix is 
re-scaled, as in \cite{DrinMuthuMahSarlos11}, and one in 
which no re-scaling takes place. In particular, let $S_{NR} = \tilde{S}$ denote the random sampling matrix (where the subscript $NR$ denotes the fact that no re-scaling takes place). The re-scaled sampling matrix is 
$S_{R} \in \mathbb{R}^{r \times n} = \tilde{S} W$, where 
$W \in \mathbb{R}^{n \times n}$ is a diagonal re-scaling matrix, where 
$[W]_{jj} = \sqrt{\frac{1}{r p_j}}$ and $W_{ji} = 0$ for $j \neq i$. 
The quantity $\frac{1}{p_j}$ is the re-scaling factor.
In this case, we have the following result, the proof of which may be found
in Section~\ref{sxn:proof-thm-one}.

\btheos
\label{ThmOne}
For $S = S_{R}$, with $r \geq \frac{C p}{(1-\theta)} \log\big(\frac{C' p}{(1-\theta)} \big)$,
then with probability at least $0.7$,
it holds that $\mbox{rank}(S_R U) = p$ and that:
\begin{eqnarray*}
C_{WC}(S_{R}) & \leq & 1+12 \frac{p}{r} \\
C_{PE}(S_{R}) & \leq & 44 \frac{n}{r}\\
C_{RE}(S_{R}) & \leq & 1+ 44 \frac{p}{r}  .
\end{eqnarray*}
\etheos
Several things are worth noting about this result.
First, note that both $C_{WC}(S_{R})-1$ and $C_{RE}(S_{R})-1$ scale as 
$\frac{p}{r}$; thus, it is possible to obtain high-quality performance 
guarantees for ordinary least squares, as long as $\frac{p}{r} \rightarrow 0$, 
e.g., if $r$ is only slightly larger than $p$.
On the other hand, $C_{PE}(S_{R})$ scales as $\frac{n}{r}$, meaning $r$ 
needs to be close to $n$ to provide similar performance guarantees. 
Next, note that all of the upper bounds apply to any data matrix $X$, 
without assuming any additional structure on $X$. 
Finally, note that when $\theta = 1$, which corresponds to sampling the rows based on $\{ q_i \}_{i=1}^n$, all the upper bounds are $\infty$. 
Our simulations also reveal that uniform sampling generally performs more 
poorly than leverage-score based approaches under the linear models we 
consider.

An important practical point is the following: the distribution $\{q_i\}_{i=1}^n$ does \emph{not} enter the results. 
This allows us to consider different distributions. 
An obvious choice is uniform, i.e., $q_i = \frac{1}{n}$ (see e.g., \cite{MMY14_ICML,MMY15}). 
Another important example is that of \emph{approximate} leverage-score sampling, as developed in \cite{DMMW12_JMLR}. 
(The running time of the main algorithm of \cite{DMMW12_JMLR} is $o(np^2)$, and thus this reduces computation compared with the use of exact leverage scores, which take $O(np^2)$ time to compute).  
Let $(\tilde{\ell_i})_{i=1}^n$ denote the approximate leverage scores developed by the procedure in \cite{DMMW12_JMLR}. Based on Theorem 2 in \cite{DMMW12_JMLR}, $|\ell_i - \tilde{\ell_i}| \leq \epsilon$ where $0 < \epsilon < 1$ for $r$ appropriately chosen. Now, using $p_i = \frac{\tilde{\ell_i}}{p}$, $p_i$ can be re-expressed as $p_i = (1-\epsilon) \frac{\ell_i}{p} + \epsilon q_i$ where $(q_i)_{i=1}^n$ is a distribution (unknown since we only have a bound on the approximate leverage scores). Hence, the performance bounds achieved by approximate leveraging are analogous to those achieved by adding $\epsilon$ multiplied by a uniform or other arbitrary distribution. 

Next, we consider the leverage-score estimator without re-scaling $S_{NR}$. In order to develop nontrivial bounds on $C_{WC}(S_{NR})$, $C_{PE}(S_{NR})$, and $C_{RE}(S_{NR})$, we need to make a strong assumption on the leverage-score distribution on $X$.
To do so, we define the following.

\bdes[k-heavy hitter leverage distribution]
A sequence of leverage scores $(\ell_i)_{i=1}^n$ is a \emph{k-heavy hitter} leverage score distribution if there exist constants $c, C > 0$ such that for $1 \leq i \leq k$, $\frac{c p}{k} \leq \ell_i \leq \frac{C p}{k}$ and for the remaining $n-k$ leverage scores, $\sum_{i=k+1}^p {\ell_i} \leq \frac{3}{4}$.
\edes
\noindent
The interpretation of a $k$-heavy hitter leverage distribution is one in which only $k$ samples in $X$ contain the majority of the leverage score mass. In the simulations below, we provide examples of synthetic matrices $X$ where the majority of the mass is in the largest leverage scores. The parameter $k$ acts as a measure of non-uniformity, in that the smaller the $k$, the more non-uniform are the leverage scores. The $k$-heavy hitter leverage distribution allows us to model highly non-uniform leverage scores.
In this case, we have the following result, the proof of which may be found
in Section~\ref{sxn:proof-thm-two}.

\btheos
\label{ThmTwo}
For $S = S_{NR}$, with $\theta = 0$ and assuming a $k$-heavy hitter leverage distribution and $r \geq c_1 p \log\big(c_2 p\big)$, 
then with probability at least $0.6$,
it holds that $\mbox{rank}(S_{NR}) = p$ and that:
\begin{eqnarray*}
C_{WC}(S_{NR}) & \leq & 1+ \frac{44 C^2}{c^2} \frac{p}{r} \\
C_{PE}(S_{NR}) & \leq & \frac{44 C^4}{c^2} \frac{k}{r}\\
C_{RE}(S_{NR}) & \leq & 1 + \frac{44 C^4}{c^2} \frac{p k}{n r}  .
\end{eqnarray*}
\etheos
Notice that when $k \ll n$, bounds in Theorem~\ref{ThmTwo} on $C_{PE}(S_{NR})$ and $C_{RE}(S_{NR})$ are significantly sharper than bounds in Theorem~\ref{ThmOne} on $C_{PE}(S_{R})$ and $C_{RE}(S_{R})$. Hence not re-scaling has the potential to provide sharper bound in the statistical setting. However a stronger assumption on $X$ is needed for this~result.

\subsection{Random projection methods}

Here, we consider two random projection algorithms, one based on a sub-Gaussian projection matrix and the other based on a Hadamard projection matrix. To do so, define $[S_{SGP}]_{ij} = \frac{1}{\sqrt{r}} X_{ij}$, where $(X_{ij})_{1\leq i \leq r, 1 \leq j \leq n}$ are i.i.d. sub-Gaussian random variables with $\mathbb{E}[X_{ij}] = 0$, variance $\mathbb{E}[X_{ij}^2] = \sigma^2$ and sub-Gaussian paramater $1$. 
In this case, we have the following result, the proof of which may be found
in Section~\ref{sxn:proof-thm-three}.

\btheos
\label{ThmThree}
For any matrix $X$, there exists a constant $c$ such that if $r \geq c' \log n$, 
then with probability greater than $0.7$, 
it holds that $\mbox{rank}(S_{SGP}) = p$ and that:
\begin{eqnarray*}
C_{WC}(S_{SGP}) & \leq & 1 + 11 \frac{p}{r}\\
C_{PE}(S_{SGP}) & \leq & 44(1 + \frac{n}{r})\\
C_{RE}(S_{SGP}) & \leq & 1 + 44 \frac{p}{r}  .
\end{eqnarray*}
\etheos

Notice that the bounds in Theorem~\ref{ThmThree} for $S_{SGP}$ are equivalent to the bounds in Theorem~\ref{ThmOne} for $S_{R}$, except that $r$ is required only to be larger than $O(\log n)$ rather than $O(p \log p)$. Hence for smaller values of $p$, random sub-Gaussian projections are more stable than leverage-score sampling based approaches. This reflects the fact that to a first-order approximation, leverage-score sampling performs as well as performing a smooth projection. 

Next, we consider the randomized Hadamard projection matrix. In particular, $S_{Had} = S_{Unif} H D$, where $H \in \mathbb{R}^{n \times n}$ is the standard Hadamard matrix (see e.g., \cite{Hedayat78}), $S_{Unif} \in \mathbb{R}^{r \times n}$ is an $r \times n$ uniform sampling matrix, and $D \in \mathbb{R}^{n \times n}$ is a diagonal matrix with random equiprobable $\pm 1$ entries.
In this case, we have the following result, the proof of which may be found
in Section~\ref{sxn:proof-thm-four}.

\btheos
\label{ThmFour}
For any matrix $X$, there exists a constant $c$ such that if $r \geq c p \log n ( \log p + \log \log n)$, 
then with probability greater than $0.8$, 
it holds that $\mbox{rank}(S_{Had}) = p$ and that:
\begin{eqnarray*}
C_{WC}(S_{Had}) & \leq & 1 + 40 \log(np) \frac{p}{r}\\
C_{RE}(S_{Had}) & \leq & 40\log (np) (1 + \frac{n}{r})\\
C_{PE}(S_{Had}) & \leq & 1 + 40\log (np) (1 + \frac{p}{r}).  
\end{eqnarray*}
\etheos
Notice that the bounds in Theorem~\ref{ThmFour} for $S_{Had}$ are equivalent to the bounds in Theorem~\ref{ThmOne} for $S_{R}$, up to a constant and $\log(np)$ factor. As discussed in \cite{DrinMuthuMahSarlos11}, the Hadamard transformation makes the leverage scores of $X$ approximately uniform (up to a $\log(np)$ factor), which is why the performance is similar to the sub-Gaussian projection (which also tends to make the leverage scores of $X$ approximately uniform). We suspect that the additional $\log(np)$ factor is an artifact of the analysis since we use an entry-wise concentration bound; using more sophisticated techniques, we believe that the $\log(np)$ can be removed.

\subsection{Lower Bounds}

Subsequent to the dissemination of the original version of this paper \cite{RaskuttiMahoney}, \cite{PilanciWainwright} amongst other results develop lower bounds on the numerator in $C_{PE}(S)$.
This proves that our upper bounds on $C_{PE}(S)$ can not be improved. We re-state Theorem 1 (Example 1) in \cite{PilanciWainwright} in a way that makes it most comparable to our results.

\btheos[Theorem 1 in \cite{PilanciWainwright}]
For any sketching matrix satisfying $\|\mathbb{E}[S^T(S S^T)^{-1}S]\|_{op} \leq \eta \frac{r}{n}$, any estimator based on $(SX, SY)$ satisfies the lower bound with probability greater than $1/2$:
\begin{eqnarray*}
C_{PE}(S) & \geq & \frac{n}{128 \eta r}.
\end{eqnarray*}
\etheos

Gaussian and Hadamard projections, as well as re-weighted approximate leverage-score sampling, all satisfy the condition $\|\mathbb{E}[S^T(S S^T)^{-1}S]\|_{op} \leq \eta \frac{r}{n}$. On the other hand, un-weighted leverage-score sampling does not satisfy this condition and hence does not satisfy the lower bound. This is why we are able to prove a tighter upper bound when the matrix $X$ has highly non-uniform leverage scores. 
Importantly, this proves that $C_{PE}(S)$ is a quantity that is more challenging to control than $C_{RE}(S)$ and $C_{WC}(S)$ when only a single sketch is used. Using this insight, \cite{PilanciWainwright} show that by using a particular iterative Hessian sketch, $C_{PE}(S)$ can be controlled up to constant.  In addition to providing a lower bound on the PE using a sketching matrix just once, \cite{PilanciWainwright} also develop a new iterative sketching scheme where sketching matrices are used repeatedly can reduce the PE significantly.

Finally, in prior work, \cite{LuFoster14,LuFoster13} show that the rate $1+\frac{p}{r}$ may be achieved for the PE using the estimator $\tilde{\beta} = ((SX)^T (SX))^{-1} X^T Y$. This estimator is related to the ridge regression estimator since sketches or random projections are applied only in the computation of the $X^T X$ matrix and not $X^T Y$. Since both $X^T Y$ and $(SX)^T (SX)$ have small dimension, this estimator has significant computational benefits. However this estimator does not violate the lower bound in \cite{PilanciWainwright} since it not based on the sketches $(SX, SY)$ but instead uses $(SX, X^T Y)$.

\section{Empirical results}
\label{SecSimulations}

\begin{figure}[h]
\begin{center}
\begin{tabular}{ccc}
{\includegraphics[scale=0.4]{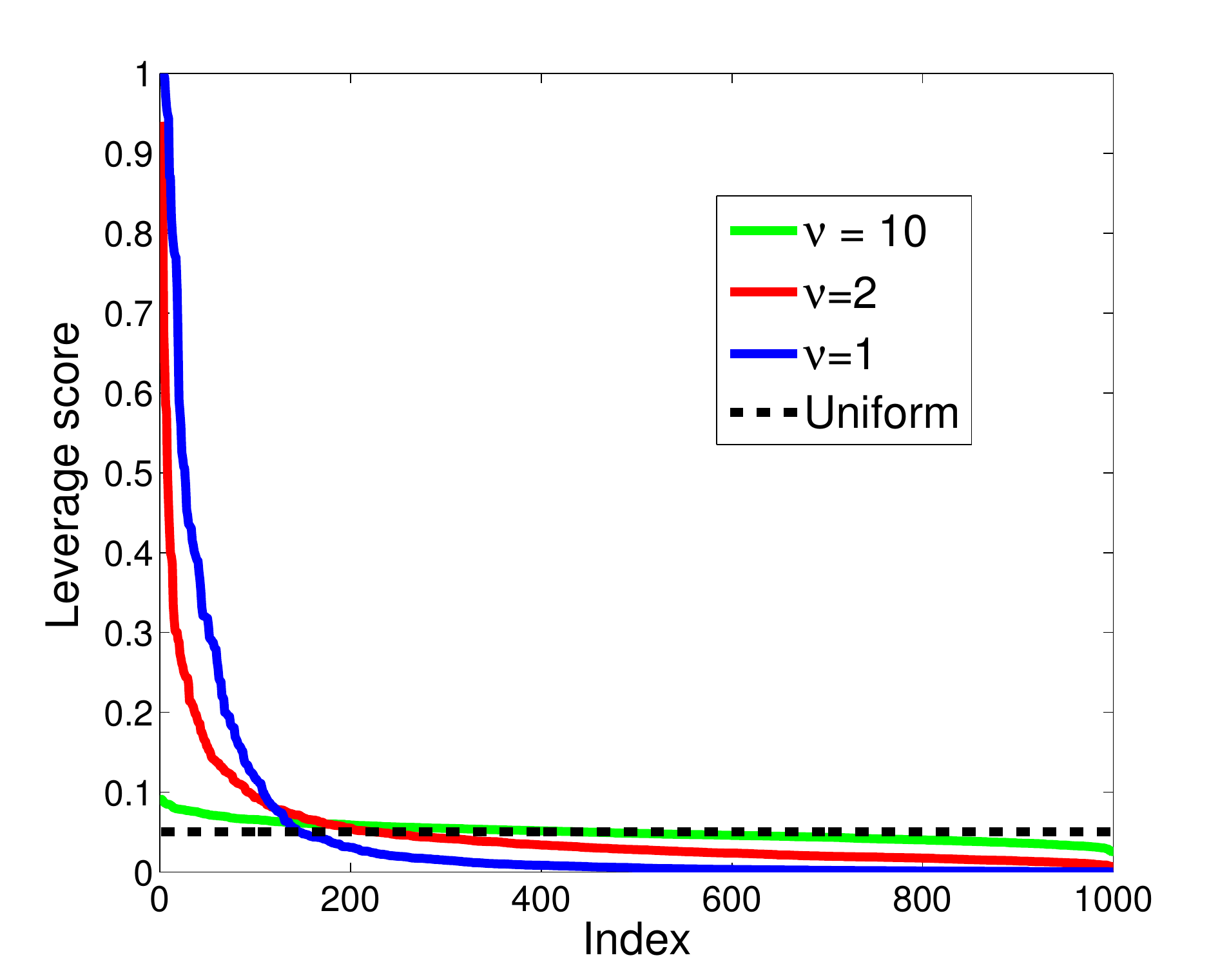}} & & 
{\includegraphics[scale=0.4]{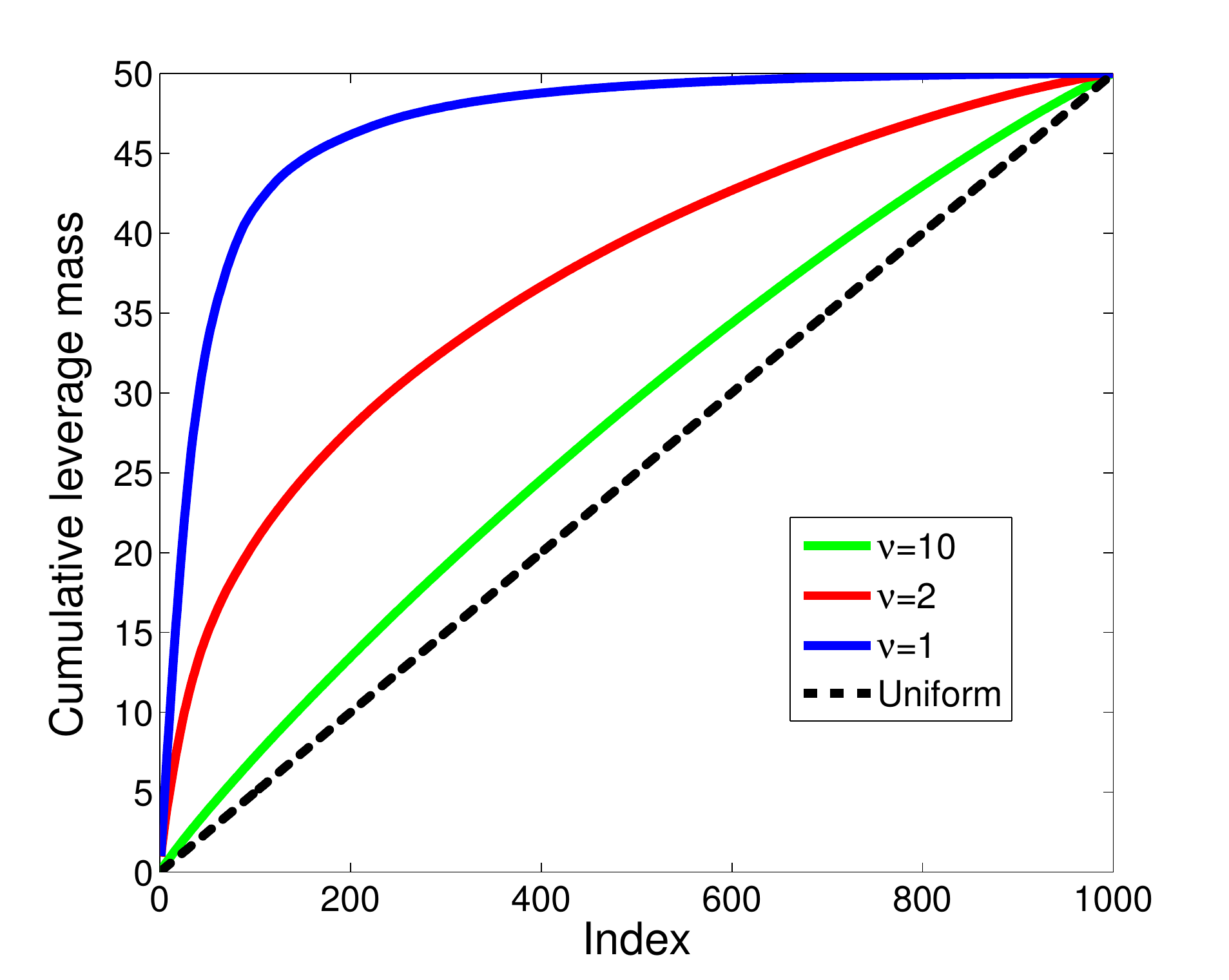}} 
\end{tabular}
\end{center}
\caption{Ordered leverage scores for different values of $\nu$ (a) and cumulative sum of ordered leverage scores  for different values of $\nu$ (b).}
\label{FigLeverage}
\end{figure}

In this section, we present the results of an empirical evaluation, illustrating the results of our theory.  We will compare the following $6$ sketching matrices.
\begin{enumerate}
\item[(1)] $S = S_{R}$ - random leverage-score sampling with re-scaling.
\item[(2)] $S = S_{NR}$ - random leverage-score sampling without re-scaling.
\item[(3)] $S=S_{Unif}$ - random uniform sampling (each sample drawn independently with probability $1/n$).
\item[(4)] $S = S_{Shr}$ - random leverage-score sampling with re-scaling and with $\theta = 0.1$.
\item[(5)] $S = S_{GP}$ - Gaussian projection matrices.
\item[(6)] $S = S_{Had}$ - Hadamard projections.
\end{enumerate}
To compare the methods and see how they perform on inputs with different leverage scores, we generate test matrices using a method outlined in \cite{MMY14_ICML,MMY15}. Set $n=1024$ (to ensure, for simplicity, an integer power of $2$ for the Hadamard transform) and $p=50$, and let the number of samples drawn with replacement, $r$, be varied. $X$ is then generated based on a t-distribution with different choices of $\nu$ to reflect different uniformity of leverage scores. Each row of $X$ is selected independently with distribution $X_i \sim t_\nu(\Sigma)$, where $\Sigma$ corresponds to an auto-regressive model with $\nu$ the degrees of freedom. The $3$ values of $\nu$ presented here are $\nu = 1$ (highly non-uniform), $\nu = 2$ (moderately non-uniform), and $\nu = 10$ (very uniform). See Figure ~\ref{FigLeverage} for a plot to see how $\nu$ influences the uniformity of the leverage scores. 
For each setting, the simulation is repeated $100$ times in order to average over both the randomness in the sampling, and in the statistical setting, the randomness over $y$.

Note that a natural comparison can be drawn between the parameter $\nu$ and the parameter $k$ in the $k$-heavy hitter definition. If we want to find the value $k$ such that $90 \%$ of the leverage mass is captured, for $\nu = 1$, $k \approx 100$, for $\nu = 2$, $k \approx 700$ and for $\nu = 10$, $k \approx 900$, according to Figure 1 (b). Hence the smaller $\nu$, the smaller $k$ since the leverage-scores are more non-uniform.

\begin{figure}[h]
\begin{center}
\begin{tabular}{ccccc}
{\includegraphics[scale = 0.23]{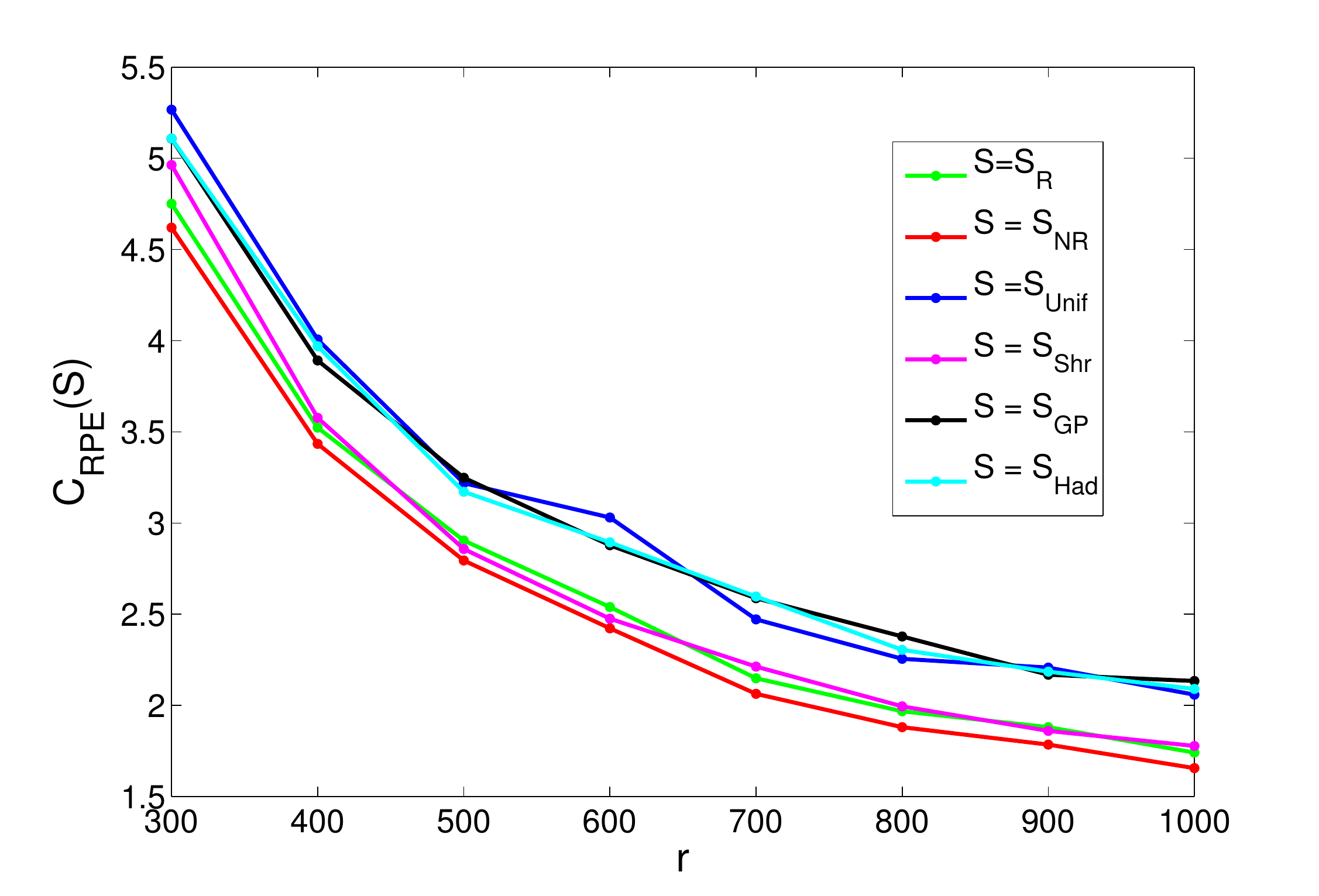}} &  
{\includegraphics[scale = 0.23]{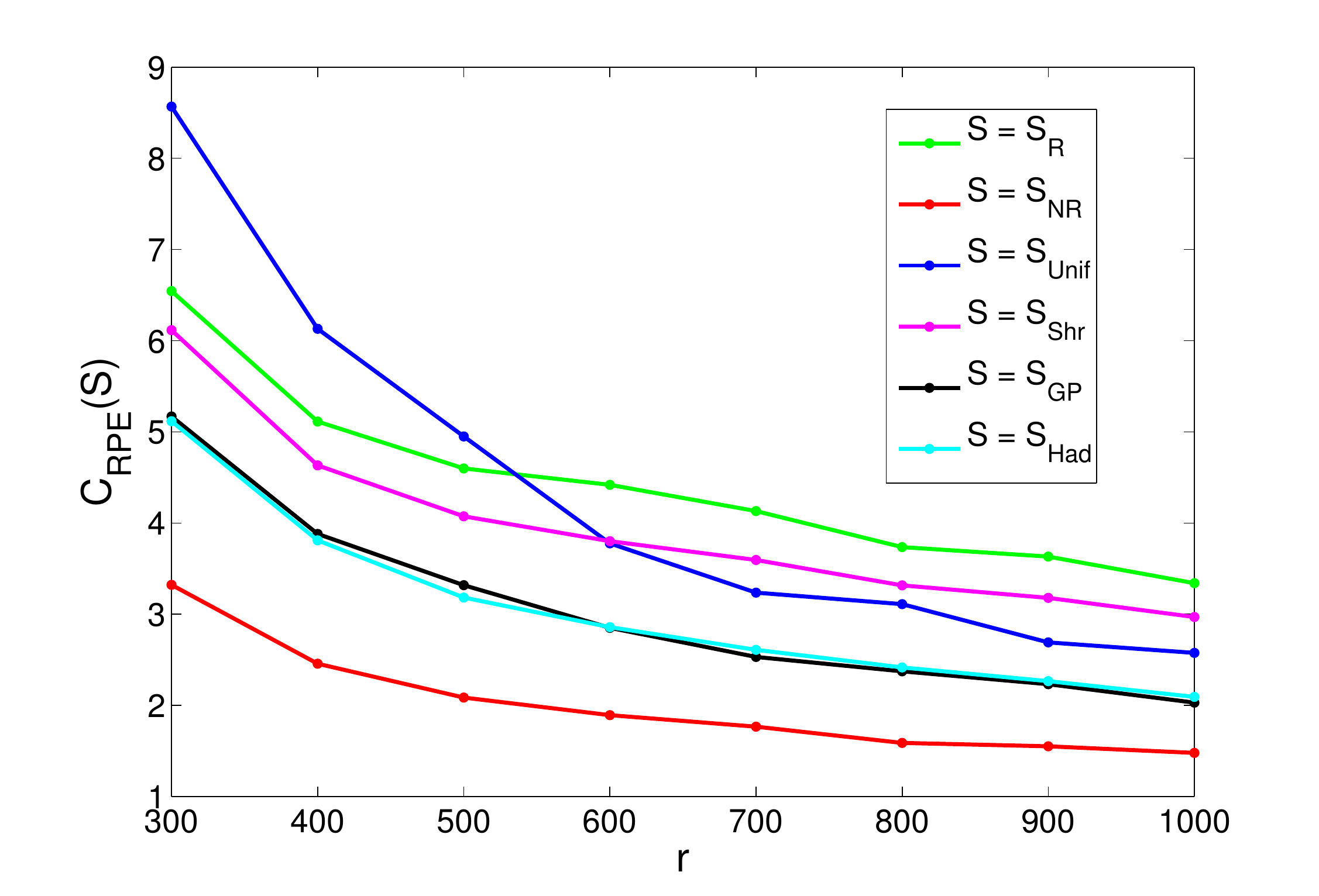}}  & 
{\includegraphics[scale = 0.23]{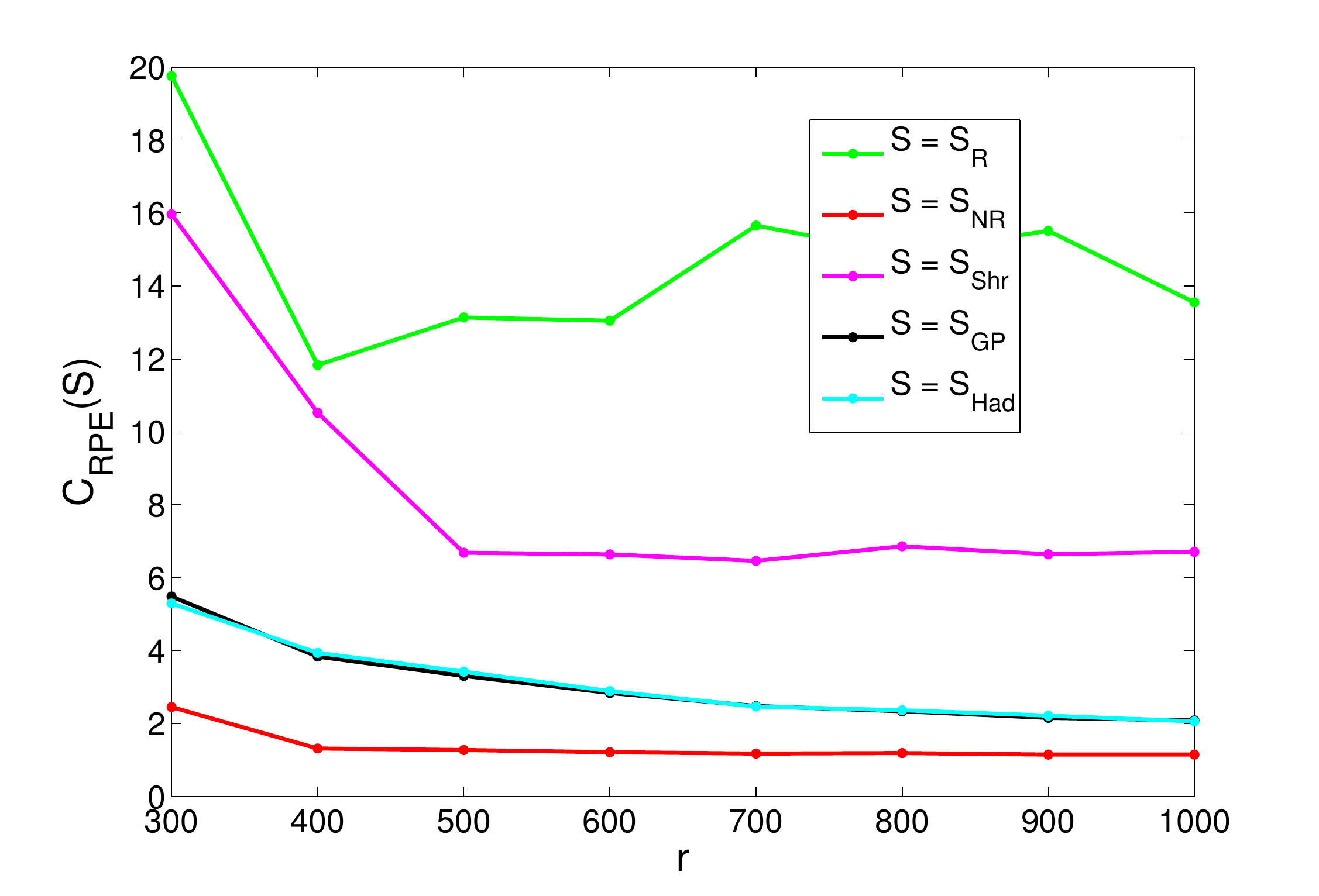}}  \\
(a) $\nu = 10$ &  (b) $\nu = 2$ &  (c) $\nu =1$
\end{tabular}
\end{center}
\caption{Relative prediction efficiency $C_{PE}(S)$ for large $r$.}
\label{FigStatLarge}
\end{figure}

We first compare the sketching methods in the statistical setting by comparing $C_{PE}(S)$. In Figure~\ref{FigStatLarge}, we plot the average $C_{PE}(S)$ for the $6$ subsampling approaches outlined above, averaged over $100$ samples for larger values of $r$ between $300$ and $1000$. In addition, in Figure~\ref{FigStatSmall}, we include a table for results on smaller values of $r$ between $80$ and $200$, to get a sense of the performance when $r$ is close to $p$.
Observe that in the large $r$ setting, $S_{NR}$ is clearly the best approach, out-performing $S_R$, especially for $\nu = 1$. For small $r$, projection-based methods work better, especially for $\nu=1$, since they tend to uniformize the leverage scores. In addition, $S_{Shr}$ is superior compared to $S_{NR}$, when $r$ is small, especially when $\nu = 1$. We do not plot $C_{RE}(S)$ as it is simply a re-scaled $C_{PE}(S)$ to Lemma 1.

\begin{figure}[h]
\centering
\subtable[$\nu=10$]{
\begin{tabular}{| l | l | l | l | l | l | l |}
\hline
r & $S_{R}$ & $S_{NR}$ & $S_{Unif}$ & $S_{Shr}$ & $S_{GP}$ & $S_{Had}$\\
\hline
80 & $39.8$ & $38.7$ & $37.3$ & $39.8$ & $35.5$ & $40.0$\\
\hline
90 & $27.8$ & $27.2$ & $27.2$ & $27.2$ & $26.1$ & $27.4$\\
\hline
100 & $22.4$ & $22.1$ & $22.7$ & $22.2$ & $21.3$ & $23.1$\\
\hline
200 & $8.33$ & $7.88$ & $8.51$ & $7.62$ & $8.20$ & $8.14$\\
\hline
\end{tabular}
}
\hspace{0.5cm}
\subtable[$\nu =2$]{
\begin{tabular}{| l | l | l | l | l | l | l |}
\hline
r & $S_{R}$ & $S_{NR}$ & $S_{Unif}$ & $S_{Shr}$ & $S_{GP}$ & $S_{Had}$\\
\hline
80 & $70.7$ & $60.1$ & $1.05 \times 10^2$ & $73.3$ & $36.5$ & $39.8$\\
\hline
90 & $45.6$ & $34.6$ & $66.2$ & $44.7$ & $26.7$ & $28.2$\\
\hline
100 & $35.1$ & $25.9$ & $52.8$ & $33.8$ & $22.1$ & $22.5$\\
\hline
200 & $9.82$ & $5.54$ & $15.3$ & $9.17$ & $7.59$ & $7.81$\\
\hline
\end{tabular}
}
\hspace{0.5cm}
\subtable[$\nu =1$]{
\begin{tabular}{| l | l | l | l | l | l | l |}
\hline
r & $S_{R}$ & $S_{NR}$ & $S_{Unif}$ & $S_{Shr}$ & $S_{GP}$ & $S_{Had}$\\
\hline
80 & $4.4 \times 10^4$ & $3.1 \times 10^4$ & $7.0 \times 10^3$ & $1.4 \times 10^4$ & $34.2$ & $40.0$\\
\hline
90 & $1.5 \times 10^4$ & $7.0 \times 10^3$ & $5.2 \times 10^3$ & $1.0 \times 10^4$ & $26.0$ & $28.7$\\
\hline
100 & $1.8 \times 10^4$ & $3.6 \times 10^3$ & $3.9 \times 10^3$ & $3.4 \times 10^3$ & $22.7$ & $24.8$\\
\hline
200 & $2.0 \times 10^2$ & $34.0$ & $5.2 \times 10^2$ & $3.6 \times 10^2$ & $7.94$ & $7.84$\\
\hline
\end{tabular}
} 
\caption{Relative prediction efficiency $C_{PE}(S)$ for small $r$.}
\label{FigStatSmall}
\end{figure}

Overall, $S_{NR}$, $S_{R}$, and $S_{Shr}$ compare very favorably to $S_{Unif}$, which is consistent with Theorem~\ref{ThmTwo}, since samples with higher leverage scores tend to reduce the mean-squared error. Furthermore, $S_{R}$ (which recall involves re-scaling) only increases the mean-squared error, which is again consistent with the theoretical results. The effects are more apparent as the leverage score distiribution is more non-uniform (i.e., for $\nu = 1$).
 
The theoretical upper bound in Theorems~\ref{ThmOne}-~\ref{ThmFour} suggests that $C_{PE}(S)$ is of the order $\frac{n}{r}$, independent of the leverage scores of $X$, for $S = S_{R}$ as well as $S = S_{Had}$ and $S_{GP}$. 
On the other hand, the simulations suggest that for highly non-uniform leverage scores, $C_{PE}(S_R)$ is higher than when the leverage scores are uniform, whereas for $S = S_{Had}$ and $S_{GP}$, the non-uniformity of the leverage scores does not significantly affect the bounds. 
The reason that $S_{Had}$ and $S_{GP}$ are not significantly affected by the leverage-score distribution is that the Hadamard and Gaussian projection has the effect of making the leverage scores of any matrix uniform~\cite{DrinMuthuMahSarlos11}.
The reason for the apparent disparity when $S = S_R$ is that the theoretical bounds use Markov's inequality which is a crude concentration bound. 
We suspect that a more refined analysis involving the bounded difference inequality would reflect that non-uniform leverage scores result in a larger $C_{PE}(S_R)$.

\begin{figure}[h]
\label{FigWCELarge}
\begin{center}
\begin{tabular}{ccccc}
{\includegraphics[scale = 0.23]{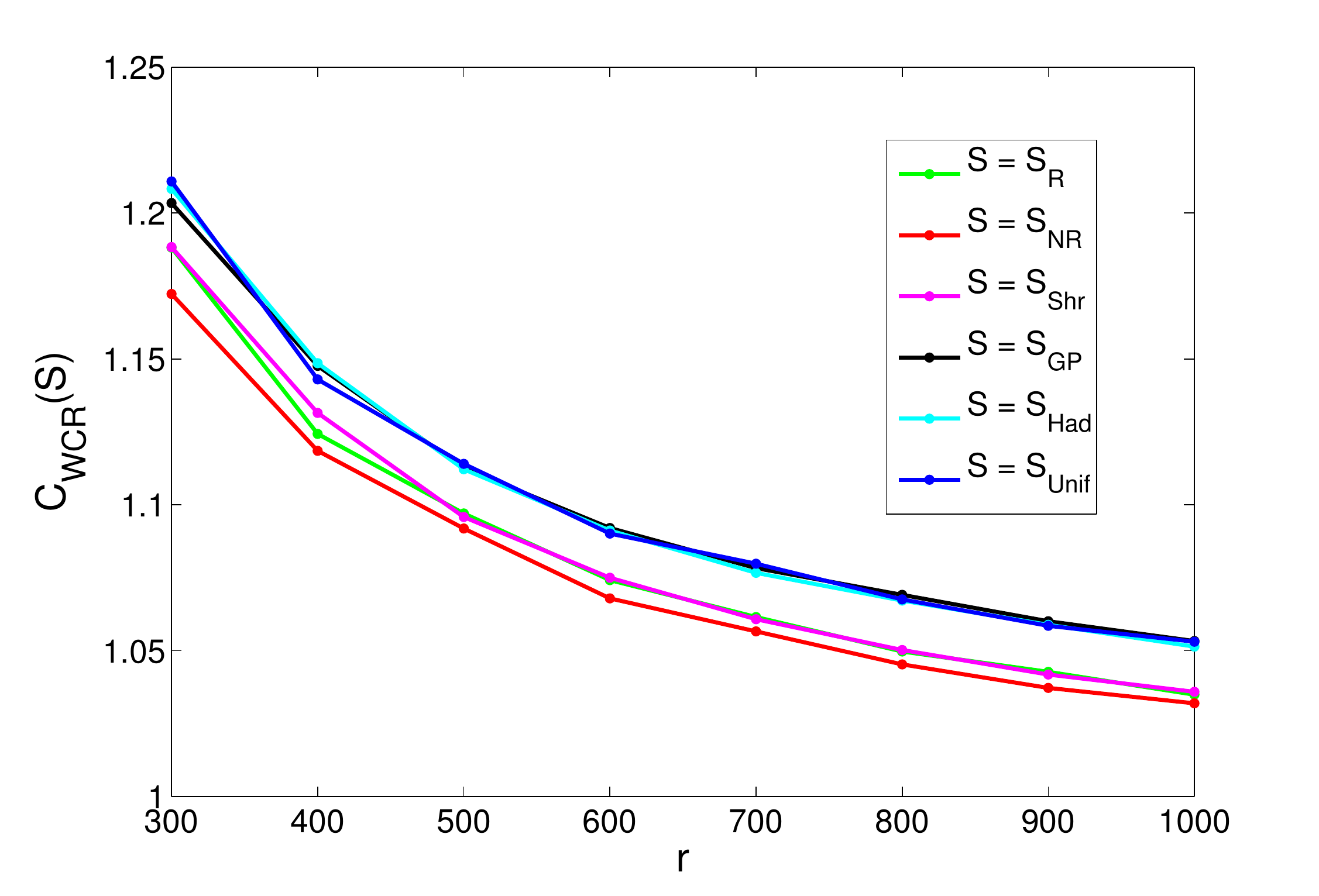}} &  
{\includegraphics[scale = 0.23]{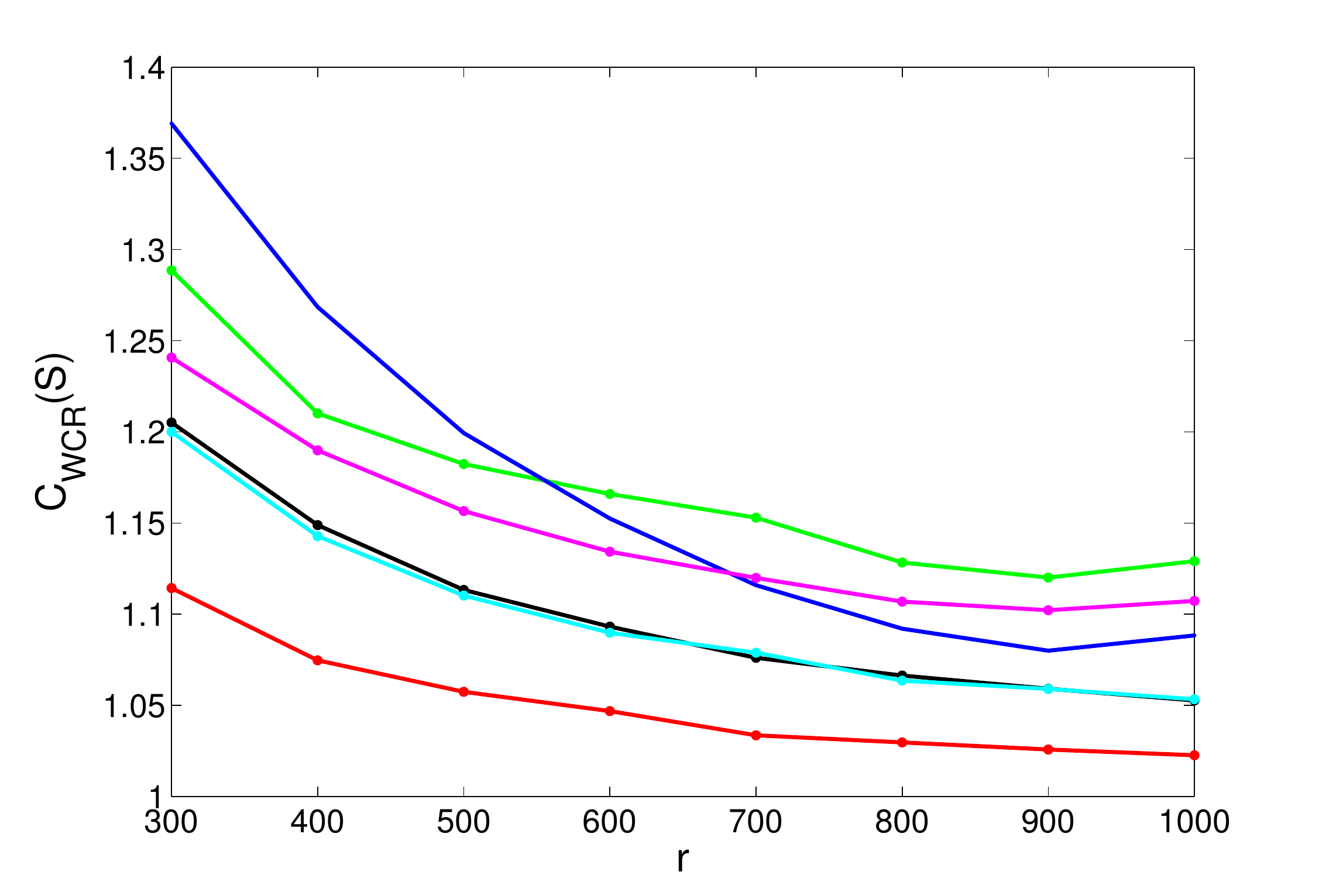}} & 
{\includegraphics[scale = 0.23]{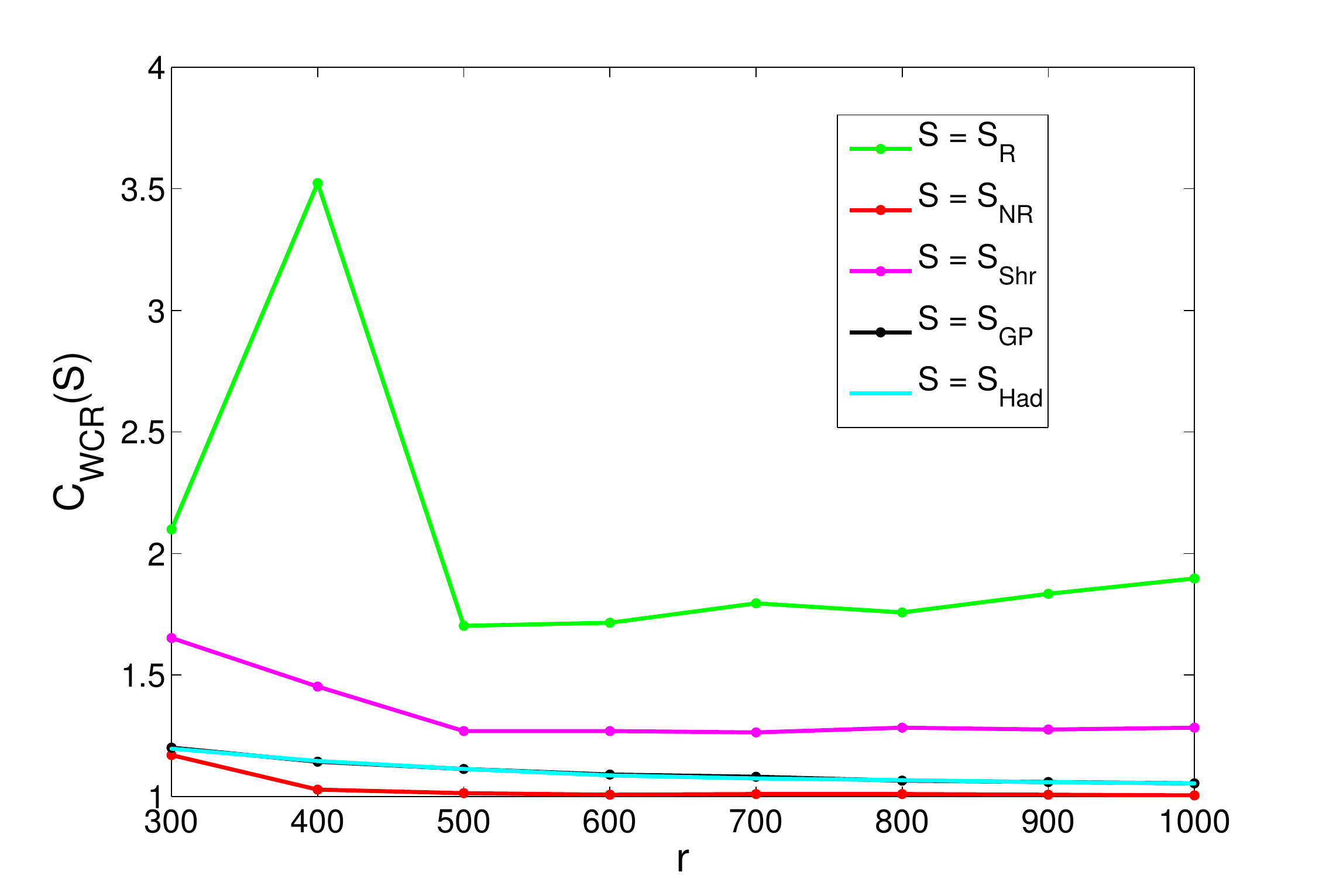}} \\
(a) $\nu = 10$ &  (b) $\nu = 2$ &  (c) $\nu =1$
\end{tabular}
\end{center}
\caption{ Worst-case relative error $C_{WC}(S)$ for large $r$.}
\end{figure}

\begin{figure}[h]
\centering
\subtable[$\nu=10$]{
\begin{tabular}{| l | l | l | l | l | l | l |}
\hline
r & $S_{R}$ & $S_{NR}$ & $S_{Unif}$ & $S_{Shr}$ & $S_{GP}$ & $S_{Had}$\\
\hline
80 & $2.82$ & $2.78$ & $2.94$ & $2.82$ & $2.74$ & $2.89$\\
\hline
90 & $2.34$ & $2.32$ & $2.40$ & $2.37$ & $2.24$ & $2.33$\\
\hline
100 & $2.04$ & $2.01$ & $2.09$ & $2.06$ & $2.02$ & $2.03$\\
\hline
200 & $1.33$ & $1.31$ & $1.36$ & $1.32$ & $1.34$ & $1.34$\\
\hline
\end{tabular}
}
\hspace{0.5cm}
\subtable[$\nu =2$]{
\begin{tabular}{| l | l | l | l | l | l | l |}
\hline
r & $S_{R}$ & $S_{NR}$ & $S_{Unif}$ & $S_{Shr}$ & $S_{GP}$ & $S_{Had}$\\
\hline
80 & $4.46$ & $3.69$ & $5.71$ & $4.33$ & $2.81$ & $2.99$\\
\hline
90 & $3.25$ & $2.85$ & $4.29$ & $3.18$ & $2.27$ & $2.28$\\
\hline
100 & $2.70$ & $2.20$ & $3.52$ & $2.61$ & $2.06$ & $2.10$\\
\hline
200 & $1.43$ & $1.22$ & $1.70$ & $1.42$ & $1.34$ & $1.36$\\
\hline
\end{tabular}
}
\hspace{0.5cm}
\subtable[$\nu =1$]{
\begin{tabular}{| l | l | l | l | l | l | l |}
\hline
r & $S_{R}$ & $S_{NR}$ & $S_{Unif}$ & $S_{Shr}$ & $S_{GP}$ & $S_{Had}$\\
\hline
80 & $6.0 \times 10^9$ & $6.0 \times 10^10$ & $2.1 \times 10^5$ & $6.9 \times 10^2$ & $2.64$ & $2.85$\\
\hline
90 & $1.7 \times 10^5$ & $3.7 \times 10^4$ & $2.4 \times 10^5$ & $5.0 \times 10^2$ & $2.35$ & $2.35$\\
\hline
100 & $9.1 \times 10^4$ & $4.5 \times 10^4$ & $1.3 \times 10^5$ & $1.8 \times 10^2$ & $2.07$ & $2.12$\\
\hline
200 & $2.1 \times 10^2$ & $70.0$ & $1.6 \times 10^4$ & $18.0$ & $1.34$ & $1.35$\\
\hline
\end{tabular}
} 
\caption{Worst-case relative error $C_{WC}(S)$ for large $r$.}
\label{FigWCESmall}
\end{figure}

Finally, Figures~\ref{FigWCELarge} and~\ref{FigWCESmall} provide a comparison of the worst-case relative error $C_{WCE}(S)$ for large and small ($r > 200$ and $r \leq 200$, respectively) values of $r$. Observe that, in general, $C_{WC}(S)$ are much closer to $1$ than $C_{PE}(S)$ for all choices of $S$. This reflects the scaling of $\frac{p}{n}$ difference between the bounds. Interestingly, Figures~\ref{FigWCELarge} and ~\ref{FigWCESmall} indicates that $S_{NR}$ still tends to out-perform $S_R$ in general, however the difference is not as significant as in the statistical setting.

\section{Discussion and conclusion}
\label{SecDiscussion}

In this paper, we developed a framework for analyzing algorithmic and statistical criteria for general sketching matrices $S \in \mathbb{R}^{r \times n}$ applied to the least-squares objective. As our analysis makes clear, our framework reveals that the algorithmic and statistical criteria depend on different properties of the oblique projection matrix $\Pi^U_S = U(SU)^{\dagger} U$, where $U$ is the left singular matrix for $X$. In particular, the algorithmic criteria (WC) depends on the quantity $\sup_{U^T \epsilon = 0} \frac{\|\Pi^U_S \epsilon \|_2}{\|\epsilon\|_2}$, since in that case the data may be arbitrary and worst-case, whereas the two statistical criteria (RE and PE) depends on $\| \Pi^U_S\|_F$, since in that case the data follow a linear model with homogenous noise variance.

Using our framework, we develop upper bounds for $3$ performance criteria applied to $4$ sketching schemes. Our upper bounds reveal that in the regime where $ p < r \ll n$, our sketching schemes achieve optimal performance up to constants, in terms of WC and RE. On the other hand, the PE scales as $\frac{n}{r}$ meaning $r$ needs to be close to (or greater than) $n$ for good performance.
Subsequent lower bounds in \cite{PilanciWainwright} show that this upper bound can not be improved, but subsequent work by \cite{PilanciWainwright} as well as \cite{LuFoster14,LuFoster13} provide alternate more sophisticated sketching approaches to deal with these challenges. 
Our simulation results reveal that for when $r$ is very close to $p$, projection-based approaches tend to out-perform sampling-based approaches since projection-based approaches tend to be more stable in that regime. 

There are numerous ways in which the framework and results from this paper can be extended. Firstly, there is a large literature that presents a number of different approaches to sketching. Since our framework provides general conditions to assess the statistical and algorithmic performance for sketching matrices, a natural and straightforward extension would be to use our framework to compare other sketching matrices. Another natural extension is to determine whether aspects of the framework can be adapted to other statistical models and problems of interest (e.g., generalized linear models, covariance estimation, PCA, etc.). Finally, another important direction is to compare the stability and robustness properties of different sketching matrices. Our current analysis assumes a known linear model, and it is unclear how the sketching matrices behave under model mis-specification.

\vspace{5mm}
\noindent
\paragraph{Acknowledgement.}
We would like to thank the Statistical and Applied Mathematical Sciences 
Institute and the members of its various working groups for helpful 
discussions.

\appendix

\section{Auxiliary Results}
\label{SecProofs}

In this section, we provide proofs of Lemma~\ref{LemProj} and an intermediate result we will later use to prove
the main theorems.  

\subsection{Proof of Lemma~\ref{LemProj}}
\label{SecProofs-first-lem}

Recall that $X = U \Sigma V^T$, where $U \in \mathbb{R}^{n \times p}$, $\Sigma \in \mathbb{R}^{p \times p}$ and $V \in \mathbb{R}^{p \times p}$ denote the left singular matrix, diagonal singular value matrix and right singular matrix respectively.

First we show that $\|Y - X \beta_{OLS}\|_2^2 = \|\epsilon\|_2^2$.
To do so, observe that
\begin{eqnarray*}
\|Y - X \beta_{OLS}\|_2^2 & = &  \| Y - U \Sigma V^T \beta_{OLS}\|_2^2 ,
\end{eqnarray*}
and set $\delta_{OLS} = \Sigma V^T \beta_{OLS}$. It follows that $\delta_{OLS} = U^T Y$. Hence
\begin{equation*}
\|Y - X \beta_{OLS}\|_2^2 = \| Y - \Pi_U Y\|_2^2,
\end{equation*}
where $\Pi_U = U U^T$. For every $Y \in \mathbb{R}^n$, there exists a unique $\delta \in \mathbb{R}^p$ and $\epsilon \in \mathbb{R}^n$ such that $U^T \epsilon = 0$ and $Y = U \delta + \epsilon$. Hence 
\begin{equation*}
\|Y - X \beta_{OLS}\|_2^2 = \|(I_{n \times n} - \Pi_U) \epsilon\|_2^2 = \| \epsilon \|_2^2,
\end{equation*}
where the final equality holds since $\Pi_U \epsilon = 0$.

\noindent Now we analyze $\| Y - X \beta_S \|_2^2$.  Observe that 
\begin{eqnarray*}
\|Y - X \beta_S\|_2^2 =  \|Y - \Pi^S_U Y \|_2^2,
\end{eqnarray*}
where $\Pi_S^U  = U(SU)^{\dagger}S$. Since $Y = U \delta + \epsilon$, it follows that
\begin{eqnarray*}
\|Y - X \beta_S\|_2^2 & = & \|U(I_{p \times p} - (SU)^{\dagger} SU ) \delta + (I_{n \times n} - \Pi^S_U) \epsilon \|_2^2\\
& = & \|(I_{p \times p} - (SU)^{\dagger} SU ) \delta\|_2^2 + \|(I_{n \times n} - \Pi^S_U) \epsilon \|_2^2\\
& = & \|(I_{p \times p} - (SU)^{\dagger} SU ) \delta\|_2^2 + \|\epsilon\|_2^2 + \|\Pi^S_U \epsilon \|_2^2  .
\end{eqnarray*}
Therefore for all $Y$:
\begin{eqnarray*}
C_{WC}(S) = \frac{\|Y - X  \beta_{S}\|_2^2}{\|Y - X \beta_{OLS}\|_2^2} = 1 + \frac{\|(I_{p \times p} - (SU)^{\dagger} SU ) \delta\|_2^2 + \|\Pi_S^U \epsilon \|_2^2}{\|\epsilon\|_2^2},
\end{eqnarray*}
where $U^T \epsilon = 0$. Taking a supremum over $Y$ and consequently over $\epsilon$ and $\delta$ completes the proof for $C_{WC}(S)$.

Now we turn to the proof for $C_{PE}(S)$. First note that
\begin{eqnarray*}
\mathbb{E}[\|X(\beta_{OLS} - \beta)\|_2^2] =  \mathbb{E}[\|U U^T Y - U \Sigma V^T \beta \| _2^2].
\end{eqnarray*}
Under the linear model $Y = U \Sigma V^T \beta + \epsilon$,
\begin{equation*}
\mathbb{E}[\|X(\beta_{OLS} - \beta)\|_2^2] =  \mathbb{E}[\|\Pi_U \epsilon \| _2^2].
\end{equation*}
Since $\mathbb{E}[\epsilon \epsilon^T] = I_{n \times n}$, it follows that
\begin{equation*}
\mathbb{E}[\|X({\beta}_{OLS} - \beta)\|_2^2] =  \mathbb{E}[\|\Pi_U \epsilon \| _2^2] = \|\Pi_U\|_F^2 = p.
\end{equation*}

\noindent For $\beta_S$, we have that
\begin{eqnarray*}
\mathbb{E}[\|X(\beta_S - \beta)\|_2^2] =  \mathbb{E}[\| \Pi_S^U Y - U \Sigma V^T \beta \|_2^2] & = & \mathbb{E}[\|(U(I- (SU)^{\dagger} SU) \Sigma V^T \beta + \Pi_S^U \epsilon \|_2^2] \\ & = & \| (I
_{p \times p}- (SU)^{\dagger} SU) \Sigma V^T \beta\|_2^2 + \mathbb{E}[\|\Pi_S^U \epsilon \|_2^2] \\
&= & \| (I
_{p \times p}- (SU)^{\dagger} SU) \Sigma V^T \beta\|_2^2 + \|\Pi_S^U \|_F^2 .
\end{eqnarray*}
Hence $C_{PE}(S) = 1/p(\| (I
_{p \times p}- (SU)^{\dagger} SU) \Sigma V^T \beta\|_2^2 +\|\Pi_S^U\|_F^2 )$ as stated.

For $C_{RE}(S)$, the mean-sqaured error for $\delta_{OLS}$ and $\delta_S$ are
\begin{eqnarray*}
\mathbb{E}[\|Y - X \beta_{OLS}\|_2^2] 
   &=&  \mathbb{E}[\|(I-\Pi^U) \epsilon \|_2^2]  \\
   &=& \| I - \Pi^U \|_F^2 = n-p ,
\end{eqnarray*}
and
\begin{eqnarray*}
\mathbb{E}[\|Y - X \beta_S\|_2^2] 
   & = & \| (I _{p \times p}- (SU)^{\dagger} SU) \Sigma V^T \beta\|_2^2+ \mathbb{E}[\|(I -\Pi_S^U) \epsilon \|_2^2]  \\
   & = & \| (I _{p \times p}- (SU)^{\dagger} SU) \Sigma V^T \beta\|_2^2+\mbox{trace}((I - \Pi_S)^T(I - \Pi_S) ) \\
   & = & \| (I _{p \times p}- (SU)^{\dagger} SU) \Sigma V^T \beta\|_2^2 + \mbox{trace}(I) - 2\mbox{trace}(\Pi_S) + \|\Pi_S\|_F^2 \\
   & = & \| (I _{p \times p}- (SU)^{\dagger} SU) \Sigma V^T \beta\|_2^2 + n - 2 p + \|\Pi_S\|_F^2 \\
   & = & \| (I _{p \times p}- (SU)^{\dagger} SU) \Sigma V^T \beta\|_2^2 + n - p + \|\Pi_S\|_F^2 - p  .
\end{eqnarray*}
Hence,
\begin{eqnarray*}
C_{RE}(S) 
  &=& \frac{n - p + \| (I _{p \times p}- (SU)^{\dagger} SU) \Sigma V^T \beta\|_2^2 +\|\Pi_S\|_F^2 - p}{n-p} \\
  &=& 1 + \frac{\| (I _{p \times p}- (SU)^{\dagger} SU) \Sigma V^T \beta\|_2^2 +\|\Pi_S\|_F^2 - p}{n-p} \\
 &=& 1 + \frac{C_{PE}(S) - 1}{n/p-1}.
\end{eqnarray*}

\subsection{Intermediate Result}
\label{SecProofs-second-lem}

In order to provide a convenient way to parameterize our upper bounds for 
$C_{WC}(S)$, $C_{PE}(S)$, and $C_{RE}(S)$, we introduce the following 
three structural conditions on $S$. 
Let $\tilde{\sigma}_{min}(A)$ denote the minimum \emph{non-zero} singular 
value of a matrix $A$. 
\begin{itemize}
\item
The first condition is that there exists an $\alpha(S)>0$ such that 
\begin{equation}
\label{CondOne}
\tilde{\sigma}_{\mbox{min}}(S U) \geq \alpha(S).
\end{equation}
\item
The second condition is that there exists a $\beta(S)$ such that
\begin{equation}
\label{CondTwoAlg}
\sup_{\epsilon,\; U^T \epsilon = 0} \frac{\| U^T S^T S \epsilon\|_2}{ \|\epsilon \|_2} \leq \beta(S).
\end{equation}
\item
The third condition is that there exists a $\gamma(S)$ such that
\begin{equation}
\label{CondTwoStat}
\|U^T S^T S\|_F \leq \gamma(S).
\end{equation}
\end{itemize}
Note that the structural conditions defined by $\alpha(S)$ and $\beta(S)$ 
have been defined previously as Eqn.~(8) and Eqn.~(9) in \cite{DrinMuthuMahSarlos11}.

Given these quantities, we can state the following lemma, the proof of which
may be found in Section~\ref{SecProofs-second-lem}.
This lemma provides upper bounds for $C_{WC}(S)$, $C_{PE}(S)$, and 
$C_{RE}(S)$ in terms of the parameters $\alpha(S)$, $\beta(S)$, and 
$\gamma(S)$.

\blems
\label{LemProjOne}
For $\alpha(S)$ and $\beta(S)$, as defined in Eqn.~\eqref{CondOne} 
and~\eqref{CondTwoAlg},
\begin{equation*}
C_{WC}(S) \leq 1+\sup_{\delta \in \mathbb{R}^p, U^T \epsilon = 0 } \frac{\| (I_{p \times p} - (SU)^{\dagger}(SU) )\delta\|_2^2}{\|\epsilon\|_2^2} + \frac{\beta^2(S)}{\alpha^4(S)} .
\end{equation*}
For $\alpha(S)$ and $\gamma(S)$, as defined in Eqn.~\eqref{CondOne} 
and~\eqref{CondTwoStat},
\begin{equation*}
C_{PE}(S) \leq \frac{\| (I
_{p \times p}- (SU)^{\dagger} SU) \Sigma V^T \beta\|_2^2}{p} + \frac{\gamma^2(S)}{\alpha^4(S)}.
\end{equation*}
Furthermore,
\begin{equation*}
C_{RE}(S) \leq 1 + \frac{p}{n} \biggr [ \frac{\| (I
_{p \times p}- (SU)^{\dagger} SU) \Sigma V^T \beta\|_2^2}{p} + \frac{\gamma^2(S)}{\alpha^4(S)} \biggr ].
\end{equation*}
\elems

\noindent
Again, the terms involving $(SU)^{\dagger} SU$ are a ``bias'' that equal 
zero for rank-preserving sketching matrices.
In addition, we emphasize that the results of Lemma~\ref{LemProj} and 
Lemma~\ref{LemProjOne} hold for arbitrary sketching matrices $S$.
In Appendix~\ref{SecMainProofs}, we bound $\alpha(S)$, $\beta(S)$ and $\gamma(S)$ for 
several different randomized sketching matrices, and this will permit us 
to obtain bounds on $C_{WC}(S)$, $C_{PE}(S)$, and $C_{RE}(S)$. For the sketching matrices we analyze, we prove that the bias term is $0$ with high probability.

\subsection{Proof of Lemma~\ref{LemProjOne}}

Note that $\Pi_S^U = U(SU)^{\dagger} S$. Let $\mbox{rank}(SU) = k < p$, and the singular value decomposition is $SU = \tilde{U} \tilde{\Sigma} \tilde{V}^T$, where $\tilde{\Sigma} \in \mathbb{R}^{k \times k}$ is a diagonal matrix with non-zero singular values of $SU$. Then, 
\begin{eqnarray*}
\frac{\| \Pi_S^U \epsilon \|_2^2}{\| \epsilon\|_2^2} = \frac{\| U (SU)^{\dagger} S \epsilon \|_2^2}{\| \epsilon\|_2^2}
& = & \frac{\| (SU)^{\dagger} S \epsilon \|_2^2}{\| \epsilon\|_2^2} \\
& = & \frac{\| \tilde{V} \tilde{\Sigma}^{-1} \tilde{U}^T S \epsilon \|_2^2}{\| \epsilon\|_2^2} \\
& = & \frac{\| \tilde{V} \tilde{\Sigma}^{-2} \tilde{V}^T \tilde{V} \tilde{\Sigma} \tilde{U}^T S \epsilon \|_2^2}{\| \epsilon\|_2^2},
\end{eqnarray*}
where we have ignored the bias term which remains unchanged. Note that $\tilde{V} \tilde{\Sigma}^{-2} \tilde{V}^T \succeq \alpha^{-2}(S) I_{p \times p}$ and $\tilde{V} \tilde{\Sigma} \tilde{U}^T = (SU)^T = U^T S^T$. Hence,
\begin{eqnarray*}
\frac{\| \Pi_S^U \epsilon \|_2^2}{\| \epsilon\|_2^2}
& = & \frac{\| \tilde{V} \tilde{\Sigma}^{-2} \tilde{V}^T \tilde{V} \tilde{\Sigma} \tilde{U}^T S \epsilon \|_2^2}{\| \epsilon\|_2^2}\\
& \leq & \frac{\|U^T S^T S \epsilon \|_2^2} {\alpha^4(S) \|\epsilon\|_2^2}\\
& = & \frac{\beta^2(S)} {\alpha^4(S) \|\epsilon\|_2^2}  ,
 \end{eqnarray*}
 and the upper bound on $C_{WC}(S)$ follows.

Similarly,
\begin{eqnarray*}
\| \Pi_S^U \|_F^2 = \|U (SU)^{\dagger} S\|_F^2 = \|(SU)^{\dagger} S\|_F^2 = \| \tilde{V} \tilde{\Sigma}^{-2} \tilde{V}^T \tilde{V} \tilde{\Sigma} \tilde{U}^T S\|_F^2 \leq \alpha^{-4}(S) \|U^T S^T S\|_F^2 
\end{eqnarray*}
and the upper bound on $C_{PE}(S)$ follows.

\section{Proof of Main Theorems}

\label{SecMainProofs}

The proof techniques for all of four theorems are similar, in that we use the intermediate result Lemma~\ref{LemProjOne} and bound the expectations of $\alpha(S)$, $\beta(S)$, and $\gamma(S)$ for each $S$, then apply Markov's inequality to develop high probability bounds. 

\subsection{Proof of Theorem~\ref{ThmOne}}
\label{sxn:proof-thm-one}

First we bound $\alpha^2(S_{R})$ by using existing results in \cite{DrinMuthuMahSarlos11}. In particular applying Theorem 4 in \cite{DrinMuthuMahSarlos11} with $\beta = 1 - \theta$, $A = U^T$, $\epsilon = \frac{1}{\sqrt{2}}$ and $\delta = 0.1$ provides the desired lower bound on $\alpha(S_R)$ and ensures that the "bias" term in Lemma~\ref{LemProjOne} is $0$ since $\mbox{rank}(S_R U) = p$. 

To upper bound $\beta(S_R)$, we first upper bound its expectation, then apply Markov's inequality. Using the result of Table 1 (second row) of \cite{DrinKannanMah06} with $\beta = 1 - \theta$:
\begin{equation*}
\mathbb{E}[ \| U^T S_{R}^T S_R \epsilon \|_2^2 ] \leq \frac{1}{(1- \theta) r} \|U^T\|_F^2 \| \epsilon \|_2^2 = \frac{p}{(1- \theta) r} \| \epsilon \|_2^2.
\end{equation*}
Applying Markov's inequality, 
\begin{equation*}
\| U^T S_{R}^T S_R \epsilon \|_2^2 \leq \frac{11 p}{(1- \theta) r} \| \epsilon \|_2^2,
\end{equation*}
with probability at least $0.9$. 

Finally we bound $\gamma(S_{R})$:
\begin{eqnarray*}
\frac{1}{p} [\|U^T S_{R}^T S_{R} \|_F^2] & = & \frac{1}{p}[\mbox{trace}(U^T (S_{R}^T S_{R})^2 U)] \\
& = & \frac{1}{p}\sum_{j=1}^p \sum_{i=1}^n \sum_{k=1}^n {U_{ij} U_{kj} [(S_{R}^T S_{R})^2]_{ki} }\\
& = & \frac{1}{p}\sum_{j=1}^p \sum_{i=1}^n {U_{ij}^2 [S_{R}^T S_{R}]_{ii}^2 }\\
& = & \frac{1}{p}\sum_{i=1}^n {\ell_i [S_{R}^T S_{R}]_{ii}^2 },
\end{eqnarray*}
where the second last equality follows since $[S_{R}^T S_{R}]_{ki}^2 = 0$ for $k \neq i$ and the final equality follows since $\ell_i = \sum_{j=1}^p {U^2_{ij}}$. First we upper bound $\mathbb{E}[\gamma(S_R)]$ and then apply Markov's inequality. Recall that $[S_R]_{ki} = \frac{1}{\sqrt{r p_i}} \sigma_{ki}$ where $\mathbb{P}(\sigma_{ki} = + 1) = p_i$.  Then,

\begin{eqnarray*}
\frac{1}{p}\sum_{i=1}^n {\ell_i \mathbb{E}([S_{R}^T S_{R}]_{ii}^2])} & = & \frac{1}{r^2 p}\sum_{i=1}^n \frac{\ell_i}{p_i^2} \sum_{m=1}^r \sum_{\ell=1}^r \mathbb{E}[\sigma_{m i}^2 \sigma_{\ell i}^2] \\
& = & \frac{1}{r^2 p}\sum_{i=1}^n \frac{\ell_i}{p_i^2} \sum_{m=1}^r \sum_{\ell=1}^r \mathbb{E}[\sigma_{m i} \sigma_{\ell i}]\\
& = &   \frac{1}{r^2 p}\sum_{i=1}^n \frac{\ell_i}{p_i^2} [(r^2 - r) p_i^2 + r p_i ]\\
& = & \frac{1}{r^2 p}\sum_{i=1}^n (\ell_i (r^2 - r) + r \frac{\ell_i}{p_i})  ]\\
& = & 1 - \frac{1}{r} + \frac{1}{r p} \sum_{i=1}^n {\frac{\ell_i}{p_i}}  .
\end{eqnarray*}
Substituting $p_i = (1 - \theta) \frac{\ell_i}{p} + \theta q_i$ completes the upper bound on $\mathbb{E}[\gamma(S_R)]$:
\begin{eqnarray*}
1 - \frac{1}{r} + \frac{1}{r p} \sum_{i=1}^n {\frac{\ell_i}{p_i}} & = & 1 - \frac{1}{r} + \frac{1}{r p} \sum_{i=1}^n {\frac{\ell_i}{(1-\theta)\frac{\ell_i}{p}  + \theta q_i}}\\
& \leq & 1 - \frac{1}{r} + \frac{1}{r} \sum_{i=1}^n {\frac{1}{1-\theta}}\\
& \leq & 1 - \frac{1}{r} + \frac{n}{(1-\theta)r} \\
& \leq & 1 + \frac{n}{(1-\theta)r}  .
\end{eqnarray*}
Using Markov's inequality, 
\begin{equation*}
\mathbb{P}\big( |\gamma(S_R) - \mathbb{E}[\gamma(S_R)]| \geq 10 \mathbb{E}[\gamma(S_R)] \big) \leq 0.1,
\end{equation*}
and consequently $\gamma(S_R) \leq 11(1 + \frac{n}{(1-\theta)r})$ with probability greater than $0.9$. The final probability of $0.7$ arises since we simultaneously require all three bounds to hold which hold with probability $0.9^3 > 0.7$. Applying Lemma~\ref{LemProjOne} in combination with our high probability bounds for $\alpha(S_R)$, $\beta(S_R)$ and $\gamma(S_R)$ completes the proof for Theorem~\ref{ThmOne}.

\subsection{Proof of Theorem~\ref{ThmTwo}}
\label{sxn:proof-thm-two}

Define $\overline{S} = \sqrt{\frac{k}{r}} S_{NR}$ where $S_{NR}$ is the sampling matrix without re-scaling. Recall the $k$-heavy hitter leverage-score assumption. Since $\sum_{i=k+1}^n {\ell_i} \leq \frac{p}{10 r}$, $\sum_{i=k+1}^n {p_i} \leq \frac{1}{10 r }$ (recall $p_i = \frac{\ell_i}{p}$). Hence the probability that a sample only contains the $k$ samples with high leverage score is:
\begin{equation*}
(1 - \frac{1}{10r})^r \geq 1 - \frac{1}{10} = 0.9.
\end{equation*}
For the remainder of the proof, we condition on the event $\mathcal{A}$ that only the rows with the $k$ largest leverage scores are selected. Let $\tilde{U} \in \mathbb{R}^{k \times p}$ be the sub-matrix of $U$ corresponding to the top $k$ leverage scores. 

Let $W = \mathbb{E}[\overline{S}^T \overline{S}] \in \mathbb{R}^{k \times k}$. Since $\frac{c}{k} \leq p_i \leq \frac{C}{k}$ for all $1 \leq i \leq k$, $c I_{k \times k} \preceq W \preceq C I_{k \times k}$. Furthermore since $\sum_{i=k+1}^n{\ell_i} \leq \frac{p}{10 r}$, $0.9 I_{p \times p} \preceq \tilde{U}^T \tilde{U} \preceq I_{p \times p}$.

First we lower bound $\alpha^2(S_{NR})$. Applying Theorem 4 in \cite{DrinMuthuMahSarlos11} with $\beta = C$, $A = \tilde{U}^T W^{1/2}$, $\epsilon = \frac{c}{2}$ and $\delta = 0.1$ ensures that as long as $r \geq c' p \log(p)$ for sufficiently large $c'$, 
\begin{equation*}
\|\tilde{U}^T W \tilde{U} - \tilde{U}^T \overline{S}^T \overline{S} \tilde{U}\|_{op} \leq \frac{c}{2C},
\end{equation*}
with probability at least 0.9. Since $\tilde{U}^T W \tilde{U} \succeq \frac{3c}{4}$, $\tilde{U}^T \overline{S}^T \overline{S} \tilde{U} \succeq \frac{c}{4}$.
Therefore with probability at least 0.9,
\begin{eqnarray*}
\alpha^2(S_{NR}) \geq \frac{c r}{4 C k}.
\end{eqnarray*}

Next we bound $\beta(S_{NR})$. Since $\overline{S} = \sqrt{\frac{k}{r}} S_{NR}$, if we condition on $\mathcal{A}$, only the leading $k$ leverage scores are selected and let $\tilde{U} \in \mathbb{R}^{k \times p}$ be the sub-matrix of $U$ corresponding to the top $k$ leverage scores. Using the result of Table 1 (second row) of \cite{DrinKannanMah06} with $\beta = 1$:
\begin{equation*}
\mathbb{E}[ \| U^T S_{NR}^T S_{NR} \epsilon \|_2^2 ]  = \frac{r^2}{k^2}\mathbb{E}[ \| U^T \overline{S}^T \overline{S} \epsilon \|_2^2 ] = \frac{r^2}{k^2}\mathbb{E}[ \| \tilde{U}^T \overline{S}^T \overline{S} \epsilon \|_2^2 ]  \leq \frac{r^2}{k^2} \| \tilde{U}^T\|_2^2 \| \epsilon \|_2^2 \leq \frac{p r^2}{k^2} \| \epsilon \|_2^2.
\end{equation*}
Applying Markov's inequality, 
\begin{equation*}
\| U^T S_{NR}^T S_{NR} \epsilon \|_2^2 \leq \frac{11 p r^2}{k^2} \| \epsilon \|_2^2   ,
\end{equation*}
with probability at least $0.9$ which completes the upper bound for $\beta(S_{NR})$. 

Finally we bound $\gamma(S_{NR})$:
\begin{eqnarray*}
\mbox{trace}(\tilde{U}^T (\overline{S}^T \overline{S})^2 \tilde{U}) / p & = & \frac{1}{p} \sum_{i=1}^n [\overline{S}^T \overline{S}]_{ii}^2 \sum_{j=1}^p {\tilde{U}_{ij}^2 }\\
& = & \frac{1}{p} \sum_{i=1}^k \ell_i [\overline{S}^T \overline{S}]_{ii}^2  \\
& \leq & \frac{C k}{r} \sum_{i=1}^k \frac{1}{r}(\sum_{m=1}^r \sigma_{mi}^2 )^2   ,
\end{eqnarray*}
where the last step follows since $\ell_i \leq \frac{C p}{k}$ for $1 \leq i \leq k$ and $0$ otherwise.
Now taking expectations:
\begin{eqnarray*}
\frac{1}{r} \mathbb{E}[\sum_{i=1}^k (\sum_{m=1}^r \sigma_{mi} )^2] & = & \frac{1}{r} \sum_{i=1}^k \sum_{\ell =1}^r \sum_{m=1}^r \mathbb{E}[\sigma_{\ell i} \sigma_{mi}] \\
& \leq & \frac{C^2}{r}\sum_{i=1}^k (\frac{r^2- r}{k^2}  + \frac{r}{k})  \\
& = & C^2(\frac{r-1}{k} + 1) \\
& \leq & C^2(\frac{r}{k} + 1).
\end{eqnarray*}
Since $S_{NR} = \sqrt{\frac{k}{r}} \overline{S}$, $\mathbb{E}[\mbox{trace}(U^T (S_{NR}^T S_{NR})^2 U)/p] \leq C^2(1 + \frac{k}{r})$. Applying Markov's inequality, 
\begin{equation*}
\mbox{trace}(U^T (S_{NR}^T S_{NR})^2 U)/p \leq 11 C^2(1 + \frac{k}{r})  ,
\end{equation*}
with probability at least $0.9$. The probability of $0.6$ arises since $0.9^4 > 0.6$. Again, using Lemma~\ref{LemProjOne} completes the proof for Theorem~\ref{ThmTwo}.

\subsection{Proof of Theorem~\ref{ThmThree}}
\label{sxn:proof-thm-three}

First we bound the smallest singular value of $S_{SGP}(U)$. Using standard results for bounds on the eigenvalues of sub-Gaussian matrices (see Proposition 2.4 in \cite{RudelsonVershynin09}), each entry of $A \in \mathbb{R}^{r \times n}$ is an i.i.d. zero-mean sub-Gaussian matrix:
\begin{equation*}
\mathbb{P}( \inf_{\|x\|_2 =1} \frac{1}{\sqrt{r}} \| A x\|_2^2 \leq \frac{1}{\sqrt{2}}) \leq n \exp( - c r).  
\end{equation*}
Hence, provided $c r \geq 2 \log n$, 
\begin{equation*}
\alpha(S_{SGP}) \geq \frac{1}{\sqrt{2}}  ,
\end{equation*} 
with probability greater than $1 - c \exp(-c' r)$.

\noindent 
Next we bound $\beta(S_{SGP})$. Since $U^T \epsilon = 0$, $\| U^T S_{SGP}^T S_{SGP} \epsilon\|_2^2 = \| U^T (S_{SGP}^T S_{SGP} - I_{n \times n} )\epsilon\|_2^2$. Therefore 
\begin{eqnarray*}
\| U^T S_{SGP}^T S_{SGP} \epsilon\|_2^2 & = & \sum_{j=1}^p \biggr( {\sum_{i=1}^n \sum_{k=1}^n  U_{ij}  (S_{SGP}^T S_{SGP} - I_{n \times n} )_{ik} \epsilon_k \biggr)^2} \\
& = & \sum_{j=1}^p \sum_{i=1}^n \sum_{k=1}^n \sum_{m=1}^n \sum_{\ell=1}^n {U_{ij} U_{mj} (S_{SGP}^T S_{SGP} - I_{n \times n} )_{ik} (S_{SGP}^T S_{SGP} - I_{n \times n} )_{m \ell} \epsilon_k \epsilon_{\ell}}. 
\end{eqnarray*}
First we bound $\mathbb{E}[\| U^T S_{SGP}^T S_{SGP} \epsilon\|_2^2]$.
\begin{equation*}
\mathbb{E}[\| U^T S_{SGP}^T S_{SGP} \epsilon\|_2^2] = \sum_{j=1}^p \sum_{i=1}^n \sum_{k=1}^n \sum_{m=1}^n \sum_{\ell=1}^n U_{ij} U_{mj} \mathbb{E}[(S_{SGP}^T S_{SGP} - I_{n \times n} )_{ik} (S_{SGP}^T S_{SGP} - I_{n \times n} )_{m \ell}] \epsilon_k \epsilon_{\ell}. 
\end{equation*}
Recall that $S_{SGP} \in \mathbb{R}^{r \times n}$, $[S_{SGP}]_{s i} = \frac{X_{s i}}{\sqrt{r}}$ where $X_{s i}$ are i.i.d. sub-Gaussian random variables with mean $0$ and sub-Gaussian parameter $1$. Hence 
\begin{eqnarray*}
\mathbb{E}[(S_{SGP}^T S_{SGP} - I_{n \times n} )_{ik} (S_{SGP}^T S_{SGP} - I_{n \times n} )_{m \ell}] & = & \frac{1}{r}(\mathbb{I}(i=k)\mathbb{I}(\ell=m) + \mathbb{E}[X_i X_j X_{\ell} X_m]),
\end{eqnarray*}
where $X_i, X_k, X_{\ell}$ and $X_m$ are i.i.d . sub-Gaussian random variables. Therefore
\begin{equation*}
\mathbb{E}[\| U^T S_{SGP}^T S_{SGP} \epsilon\|_2^2] = \frac{1}{r}\sum_{j=1}^p \sum_{i=1}^n \sum_{k=1}^n \sum_{m=1}^n \sum_{\ell=1}^n U_{ij} U_{mj} (\mathbb{I}(i=k)\mathbb{I}(\ell=m) + \mathbb{E}[X_i X_k X_{\ell} X_m]) \epsilon_k \epsilon_{\ell}. 
\end{equation*}
First note that $\mathbb{I}(i=k)\mathbb{I}(\ell=m) + \mathbb{E}[X_i X_k X_{\ell} X_m] = 0$ unless $i=k$ and $\ell=m$, or $i=\ell$ and $k = \ell$ or any other combination of two pairs of variables have the same index. When $i = k = \ell = m$, $\mathbb{I}(i=k)\mathbb{I}(\ell=m) + \mathbb{E}[X_i X_k X_{\ell} X_m] = 1 + E[X_i^4] \leq 2$, since for sub-Gaussian random variables with parameter $1$, $E[X_i^4] \leq 1$ and
\begin{eqnarray*}
\frac{1}{r}\sum_{j=1}^p \sum_{i=1}^n \sum_{k=1}^n \sum_{m=1}^n \sum_{\ell=1}^n U_{ij} U_{mj} (\mathbb{I}(i=k)\mathbb{I}(\ell=m) + \mathbb{E}[X_i X_k X_{\ell} X_m]) \epsilon_k \epsilon_{\ell} & = & \frac{1}{r}\sum_{j=1}^p \sum_{i=1}^n U_{ij}^2 \epsilon_i^2(1 + E[X_i^4])\\
& \leq & \frac{2}{r}\sum_{j=1}^p \sum_{i=1}^n U_{ij}^2 \epsilon_i^2\\
& \leq & \frac{2}{r}\sum_{j=1}^p U_{ij}^2 \|\epsilon\|_2^2\\
& = & \frac{2p}{r} \|\epsilon\|_2^2.
\end{eqnarray*}
When $i=k$ and $\ell = m$ but $k \neq \ell$, $\mathbb{I}(i=k)\mathbb{I}(\ell=m) + \mathbb{E}[X_i X_k X_{\ell} X_m] = 2$ and
\begin{equation*}
\sum_{j=1}^p \sum_{i=1}^n \sum_{k=1}^n \sum_{m=1}^n \sum_{\ell=1}^n U_{ij} U_{mj} (\mathbb{I}(i=k)\mathbb{I}(\ell=m) + \mathbb{E}[X_i X_k X_{\ell} X_m]) \epsilon_k \epsilon_{\ell} = 2 \sum_{j=1}^p \sum_{i=1}^n \sum_{\ell=1}^n {U_{ij} U_{mj} \epsilon_i \epsilon_m} = 0,
\end{equation*}
since $U^T \epsilon = 0$. Using similar logic when the two pairs of variables are not identical, the sum is $0$, and hence
\begin{equation*}
\mathbb{E}[\| U^T S_{SGP}^T S_{SGP} \epsilon\|_2^2] \leq \frac{2p}{r} \|\epsilon\|_2^2, 
\end{equation*}
for all $\epsilon$ such that $U^T \epsilon = 0$. Applying Markov's inequality, 
\begin{equation*}
\| U^T S_{SGP}^T S_{SGP} \epsilon\|_2^2 \leq \frac{22p}{r} \|\epsilon\|_2^2, 
\end{equation*}
with probability greater than $0.9$. Therefore $\beta(S_{SGP}) \leq \sqrt{\frac{22p}{r} }$ with probability at least $0.9$.

Now we bound $\gamma(S_{SGP}) = \|U^T S_{SGP}^T S_{SGP} \|_F^2 = \mbox{trace}(U^T (S_{SGP}^T S_{SGP})^2 U) / p$:
\begin{eqnarray*}
\mbox{trace}(U^T (S_{SGP}^T S_{SGP})^2 U) / p & = & \frac{1}{p} \sum_{j=1}^p  \sum_{i=1}^n \sum_{k=1}^n [(S_{SGP}^T S_{SGP})^2]_{ik} {U_{ij} U_{kj} }\\
& = & \frac{1}{p r^2} \sum_{j=1}^p  \sum_{i=1}^n \sum_{k=1}^n \sum_{v=1}^n \sum_{m=1}^r \sum_{\ell = 1}^r {U_{ij} U_{kj} X_{mv}X_{mi} X_{\ell v} X_{\ell k}}  .
\end{eqnarray*}
First we bound the expectation:
\begin{eqnarray*}
\mathbb{E}[\mbox{trace}(U^T (S_{SGP}^T S_{SGP})^2 U) / p] & = & \frac{1}{p r^2} \sum_{j=1}^p  \sum_{i=1}^n \sum_{k=1}^n \sum_{v=1}^n \sum_{m=1}^r \sum_{\ell = 1}^r E[X_{mv}X_{mi} X_{\ell v} X_{\ell k}] \\
& = & \frac{1}{p r^2} \sum_{j=1}^p  \sum_{i=1}^n \sum_{k=1}^n \sum_{v=1}^n {(r^2 - r) U_{ij} U_{kj} \mathbb{E}[X_v X_i] \mathbb{E}[X_v X_k] + r \mathbb{E}[X_v^2 X_i X_k] }\\
& = & \frac{1}{p r^2} \sum_{j=1}^p  \sum_{i=1}^n {(r^2 - r) U_{ij}^2  (\mathbb{E}[X_i^2])^2} + \frac{1}{p r^2} \sum_{j=1}^p  \sum_{i=1}^n \sum_{v=1}^n {r U_{ij}^2 \mathbb{E}[X_v^2 X_i^2]}\\
& = & \frac{1}{p r^2} \sum_{j=1}^p  \sum_{i=1}^n {(r^2 - r) U_{ij}^2 \sigma^4} + \frac{1}{p r^2} \sum_{j=1}^p  \sum_{i=1}^n \sum_{v=1}^n {r U_{ij}^2 \mathbb{E}[X_v^2 X_i^2]}\\
& = & 1 - \frac{1}{r} + \frac{\mu_4}{\sigma^4 r} + \frac{n-1}{r}\\
& \leq & 1 + \frac{3}{r} + \frac{n-2}{r}\\
& = & 1 + \frac{(n+ 1)}{r}   .
\end{eqnarray*}
Applying Markov's inequality,
\begin{equation*}
\gamma(S_{SGP}) \leq 11(1 + \frac{(n+ 1)}{r}),
\end{equation*}
with probability at least $0.9$. This completes the proof for Theorem~\ref{ThmThree}.

\subsection{Proof of Theorem~\ref{ThmFour}}
\label{sxn:proof-thm-four}

For $S = S_{Had}$ we use existing results in \cite{DrinMuthuMahSarlos11} to lower bound $\alpha^2(S_{Had})$ and $\beta(S_{Had})$ and then upper bound $\gamma(S_{Had})$. Using Lemma 4 in \cite{DrinMuthuMahSarlos11} provides the desired lower bound on $\alpha^2(S_{Had})$. 

To upper bound $\beta(S_{Had})$, we use Lemma 5 in \cite{DrinMuthuMahSarlos11} which states that:
\begin{equation*}
\| U^T S_{Had}^T S_{Had} \epsilon \|_2^2 \leq \frac{20 d \log(40nd) \|\epsilon \|_2^2}{r}  ,
\end{equation*}
with probability at least $0.9$.

Finally to bound $\gamma(S_{Had})$ recall that $S_{Had} = S_{Unif} H D$, where $S_{Unif}$ is the uniform sampling matrix, $D$ is a diagonal matrix with $\pm 1$ entries and $H$ is the Hadamard matrix:
\begin{eqnarray*}
\frac{1}{p} [\|U^T S_{Had}^T S_{Had} \|_F^2] & = & \frac{1}{p}[\mbox{trace}(U^T (S_{Had}^T S_{Had})^2 U)] \\
& = & \frac{1}{p}[\mbox{trace}(U^T (D^T H^T S_{Unif}^T S_{Unif} H D)^2 U)] \\
& = & \frac{1}{p}[\mbox{trace}(U^T D^T H^T S_{Unif}^T S_{Unif} H D D^T H^T S_{Unif}^T S_{Unif} H D U)] \\
& = & \frac{1}{p}[\mbox{trace}(U^T D^T H^T (S_{Unif}^T S_{Unif})^2 H D U)] \\
& = & \frac{1}{p}\sum_{j=1}^p \sum_{i=1}^n \sum_{k=1}^n {[HDU]_{ij} [HDU]_{kj} [(S_{Unif}^T S_{Unif})^2]_{ki} } \\
& = & \frac{1}{p}\sum_{j=1}^p \sum_{i=1}^n {[HDU]_{ij}^2 [S_{Unif}^T S_{Unif}]_{ii}^2 }.
\end{eqnarray*}
Using Lemma 3 in \cite{DrinMuthuMahSarlos11}, with probability greater than $0.95$,
\begin{equation*}
\frac{1}{p}\sum_{j=1}^p {[HDU]_{ij}^2} \leq \frac{2 \log(40 n p)}{n}.
\end{equation*}
In addition, we have that
\begin{eqnarray*}
\frac{1}{p} [\|U^T S_{Had}^T S_{Had} \|_F^2] & = & \frac{1}{p}\sum_{j=1}^p \sum_{i=1}^n {[HDU]_{ij}^2 [S_{Unif}^T S_{Unif}]_{ii}^2 }\\
& \leq & \frac{2 \log(40 n p)}{n} \sum_{i=1}^n {[S_{Unif}^T S_{Unif}]_{ii}^2 }.
\end{eqnarray*}
Now we bound $\mathbb{E}[\sum_{i=1}^n {[S_{Unif}^T S_{Unif}]_{ii}^2 }]$.

\begin{eqnarray*}
\sum_{i=1}^n {\mathbb{E}([S_{Unif}^T S_{Unif}]_{ii}^2])} & = & \frac{n^2}{r^2}\sum_{i=1}^n \sum_{m=1}^r \sum_{\ell=1}^r \mathbb{E}[\sigma_{m i}^2 \sigma_{\ell i}^2] \\
& = & \frac{n^2}{r^2}\sum_{i=1}^n \sum_{m=1}^r \sum_{\ell=1}^r \mathbb{E}[\sigma_{m i} \sigma_{\ell i}]\\
& = &   \frac{n^2}{r^2}\sum_{i=1}^n  [\frac{r^2 - r}{n^2} + \frac{r}{n}]\\
& = & \frac{n^2}{r^2}[\frac{r^2 - r}{n} + r]\\
& = & n - \frac{n}{r} + \frac{n^2}{r}\\
& \leq & n + \frac{n^2}{r}.
\end{eqnarray*}

\noindent Using Markov's inequality, with probability greater than $0.9$,
\begin{equation*}
\sum_{i=1}^n {[S_{Unif}^T S_{Unif}]_{ii}^2 } \leq 10(n + \frac{n^2}{r})  ,
\end{equation*}
which completes the proof.

\vskip 0.2in
\bibliographystyle{plain} 
\bibliography{PaperFinal}

\end{document}